\documentclass[10pt, letter, onecolumn]{arxiv}

\usepackage{kantlipsum, lipsum}
\usepackage{dm-colors}
\usepackage{amsmath}
\usepackage{pstricks, pst-node}
\usepackage{verbatim}
\usepackage{multirow}
\usepackage{scalerel}
\usepackage{booktabs}
\usepackage{enumitem}
\usepackage{xspace}
\usepackage{bm}
\usepackage{bbm}
\usepackage{mathtools}
\usepackage{soul}
\usepackage{epsfig}
\usepackage{graphicx}
\usepackage{amssymb}
\usepackage{colortbl}
\usepackage{csquotes}
\usepackage{setspace}
\usepackage{tabularx,ragged2e}
\usepackage{placeins}
\usepackage{threeparttable}
\usepackage[symbol]{footmisc}
\usepackage{nameref}
\usepackage{varioref}
\usepackage[pagebackref=false,breaklinks=false,%
            colorlinks=true,bookmarks=true,citecolor=ourdarkblue,%
            urlcolor=ourdarkblue,linkcolor=ourdarkblue]{hyperref}
\usepackage[noabbrev,capitalize]{cleveref}
\usepackage{etoc}
\usepackage{lineno}

\graphicspath{{figures/}}

\newcommand{\change}[1]{{{#1}}}
\newcommand{\changee}[1]{{\textcolor{black}{#1}}}
\newcommand{\changeee}[1]{{\textcolor{black}{#1}}}

\title{\Large{PMC-VQA: Visual Instruction Tuning for Medical Visual Question Answering}}

\vspace{1cm}
\author[$\ast$,1,2]{Xiaoman Zhang}
\author[$\ast$,1,2]{Chaoyi Wu} 
\author[1,2]{Ziheng Zhao} 
\author[1,2]{Weixiong Lin} 
\author[1,2]{\\ \vspace{0.1cm} Ya Zhang}
\author[1,2,\dag]{Yanfeng Wang} 
\author[1,2,\dag]{Weidi Xie}

\affil[1]{\normalsize Shanghai Jiao Tong University \hspace{1cm}}

\affil[2]{\normalsize Shanghai AI Laboratory \authorcr 

}

\renewcommand{\correspondingauthor}[1]{$\ast$~Equal contributions. Email addresses: \{wtzxxxwcy02, xm99sjtu, weidi\}@sjtu.edu.cn \\ $\dag$~Corresponding author. }

\begin{document}


\begin{abstract}
Medical Visual Question Answering (MedVQA) presents a significant opportunity to enhance diagnostic accuracy and healthcare delivery by leveraging artificial intelligence to interpret and answer questions based on medical images.
In this study, we reframe the problem of MedVQA as a generation task that naturally follows the human-machine interaction and propose a generative-based model for medical visual understanding by aligning visual information from a pre-trained vision encoder with a large language model. 
We establish a scalable pipeline to construct a large-scale medical visual question-answering dataset, named PMC-VQA, which contains 227k VQA pairs of 149k images that cover various modalities or diseases. 
We train the proposed model on PMC-VQA and then fine-tune it on multiple public benchmarks, {\em e.g.,} VQA-RAD, SLAKE, and Image-Clef-2019, significantly outperforming existing MedVQA models in generating relevant, accurate free-form answers. 
In addition, we propose a test set that has undergone manual verification, which is significantly more challenging, serving to better monitor the development of generative MedVQA methods.
To facilitate comprehensive evaluation and comparison, we have maintained a leaderboard at \url{https://paperswithcode.com/paper/pmc-vqa-visual-instruction-tuning-for-medical}, offering a centralized resource for tracking progress and benchmarking state-of-the-art approaches.
The PMC-VQA dataset emerges as a vital resource for the field of research, and the MedVInT presents a significant breakthrough in the area of MedVQA.
\end{abstract}

\maketitle


\section{Introduction}
Large language models~(LLMs), such as GPT-4~\cite{openai2023gpt4}, Med-PaLM~\cite{singhal2022large}, PMC-LLaMA~\cite{wu2023pmcllama} have recently achieved remarkable success in the field of medical natural language processing~~\cite{jin2021disease, kung2023performance, nori2023capabilities}. 
While recent LLMs excel in language understanding in the medical domain, they are essentially ``blind'' to visual modalities, such as images and videos, hindering the use of visual content as inputs. 
This limitation is particularly evident in the Medical Visual Question Answering (MedVQA) domain, where there is a critical need for models to interpret medical visual content to answer text-based queries accurately~\cite{lin2022medical}.


\changeee{MedVQA is an important and emerging field at the intersection of artificial intelligence and healthcare, which involves developing systems that can understand and interpret medical images and provide relevant answers to questions posed about these images.
By integrating AI with medical expertise, MedVQA aims to significantly impact healthcare outcomes, patient care, and medical science~\cite{yang2023impact,thirunavukarasu2023large}.
For example, the MedVQA system can enhance diagnostic accuracy for clinicians, improve patient understanding of medical information, and advance medical education and research.}

However, existing MedVQA methods~\cite{nguyen2019overcoming,liu2021contrastive,chen2022multi,lin2023pmcclip} typically treat the problem as a retrieval task with a limited answer base and train multi-modal vision-language models with contrastive or classification objectives. Consequently, they are only useful for limited use cases where a finite set of outcomes is provided beforehand. 
We propose to develop the {\em first} open-ended MedVQA system with a generative model as the backend, capable of handling diverse questions that arise in clinical practice, generating answers in free form without being constrained by the vocabulary. 
While there has been promising research in visual-language representation learning, such as Flamingo~\cite{alayrac2022flamingo} and BLIP~\cite{li2023blip}, these models have primarily been trained on natural language and images,
with very limited application in the medical domain, 
due to the complex and nuanced visual concepts often found in medical scenarios.

\begin{figure}[tb]
    \centering
    \includegraphics[width=1\textwidth]{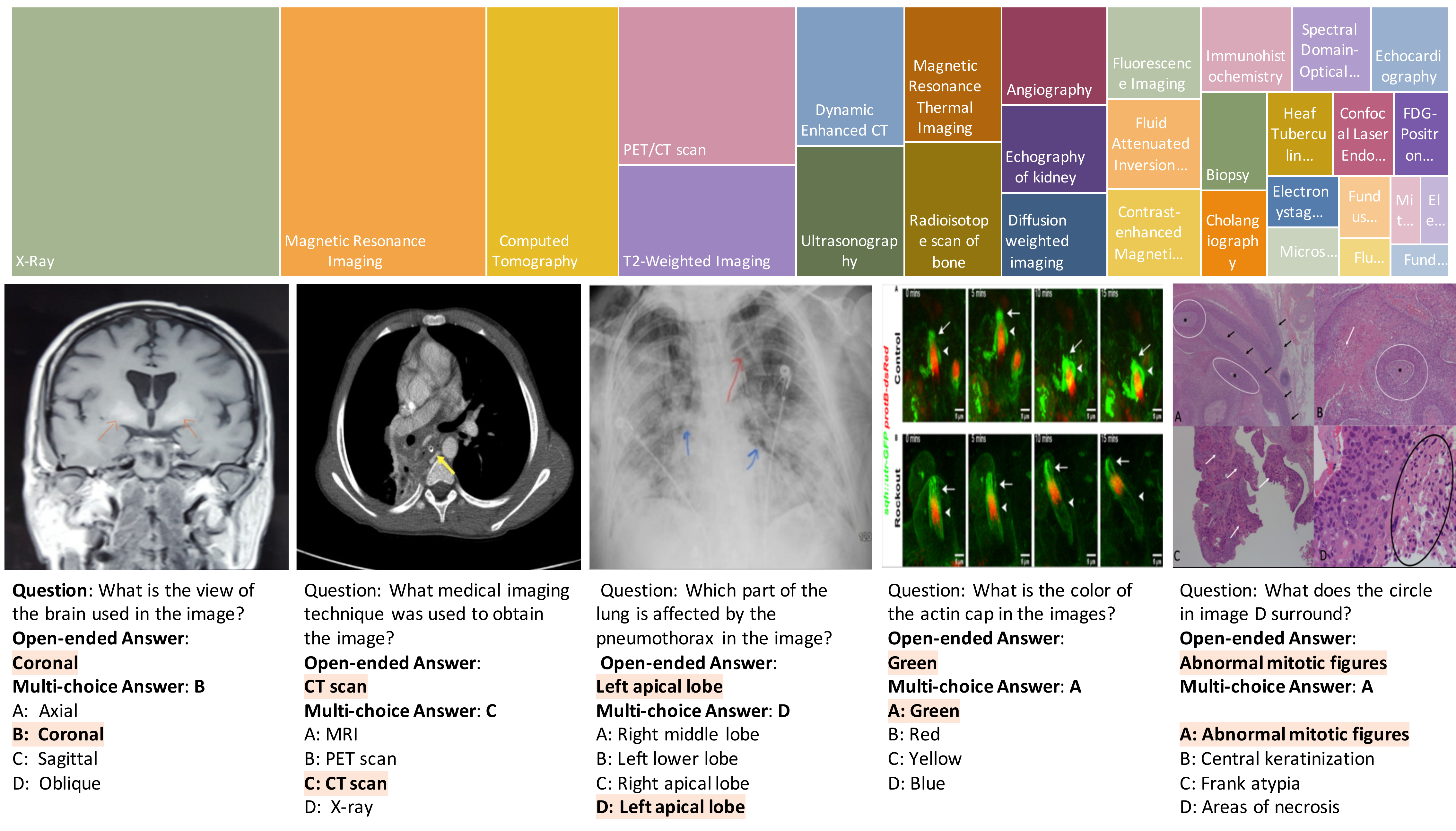}
    \caption{
    (a) Several examples of challenging questions and answers along with their respective images. 
    To answer questions related to these images, 
    the network must acquire sufficient medical knowledge, 
    for example, for the first two images, it is essential to recognize the anatomy structure and modalities;
    for the third image, recognizing the X-ray image pattern of pathologies is necessary;
    for the final two images, apart from the basic biomedical knowledge, 
    the model is also required to discern colors, differentiate subfigures, 
    and perform Optical Character Recognition (OCR).
    (b) The top 20 figure types in PMC-VQA, cover a wide range of diagnostic procedures.}
    \label{fig:image_distribution}
\end{figure}

To effectively train the generative-based models, our study reveals that existing datasets are limited in size, 
making them insufficient for training high-performing models.
we leverage well-established medical visual-language datasets~\cite{lin2023pmcclip} and initiate a scalable, automatic pipeline for constructing a new large-scale medical visual question-answering dataset. This new dataset, termed as {\bf PMC-VQA}, contains 227k VQA pairs of 149k images, 
including 80\% of radiological images,
covering various modalities or diseases~(Figure~\ref{fig:image_distribution}), surpassing existing datasets in terms of both amount and diversity.

In our experiments, we trained a generative visual-language model, termed as MedVInT, on the training set of PMC-VQA and fine-tuned it on the existing public benchmarks,
{\em e.g.}, VQA-RAD~\cite{lau2018dataset}, SLAKE~\cite{liu2021slake}, and ImageClef-VQA-2019~\cite{ben2019vqa}.
outperforming existing models by a large margin, achieving over 80\% accuracy on multi-choice selection. 
However, while evaluating our proposed challenging benchmark, 
even the state-of-the-art models struggle, showing that there is still ample room for development in this field.

In summary, our contributions are as follows:
\textbf{(i)} We reframe the problem of MedVQA as a generative learning task and propose MedVInT, a model obtained by aligning a pre-trained vision encoder with a large language model through visual instruction tuning; 
\textbf{(ii)} We introduce a scalable pipeline and construct a large-scale MedVQA dataset, PMC-VQA, which far exceeds the size and diversity of existing datasets, covering various modalities and diseases;
\textbf{(iii)} We pre-train MedVInT on PMC-VQA and fine-tune it on VQA-RAD~\cite{lau2018dataset} and SLAKE~\cite{liu2021slake}, achieving state-of-the-art performance and significantly outperforming existing models;
\textbf{(iv)} We propose a new test set and present a more challenging benchmark for MedVQA, 
to evaluate the performance of VQA methods thoroughly.

\section{Results}

The goal of our proposed model, Medical Visual Instruction Tuning~(MedVInT), is to perform generative-based medical visual question answering~(MedVQA). 
Serving for this purpose, we curate a new large-scale medical visual instruction tuning dataset, namely PMC-VQA.
In this section, 
we start by a comprehensive analysis on the PMC-VQA dataset, which contains 227k VQA pairs of 149k images, 
covering various modalities or diseases and compare it with the existing medical VQA datasts. 
Then, we will evaluate our trained model on three external MedVQA benchmarks, VQA-RAD~\cite{lau2018dataset}, SLAKE~\cite{liu2021slake} and ImageClef-VQA-2019~\cite{ben2019vqa}. Note that, our model has two variants, which are tailored to encoder-based and decoder-based language models, respectively, denoted as MedVInT-TE and MedVInT-TD. At last, we establish a novel generative MedVQA benchmark with PMC-VQA, and evaluate various pre-trained visual or language models using our framework, serving as a reference to promote the future research in generative medical VQA.


\begin{figure}[t]
    \centering
    \includegraphics[width=0.98\textwidth]{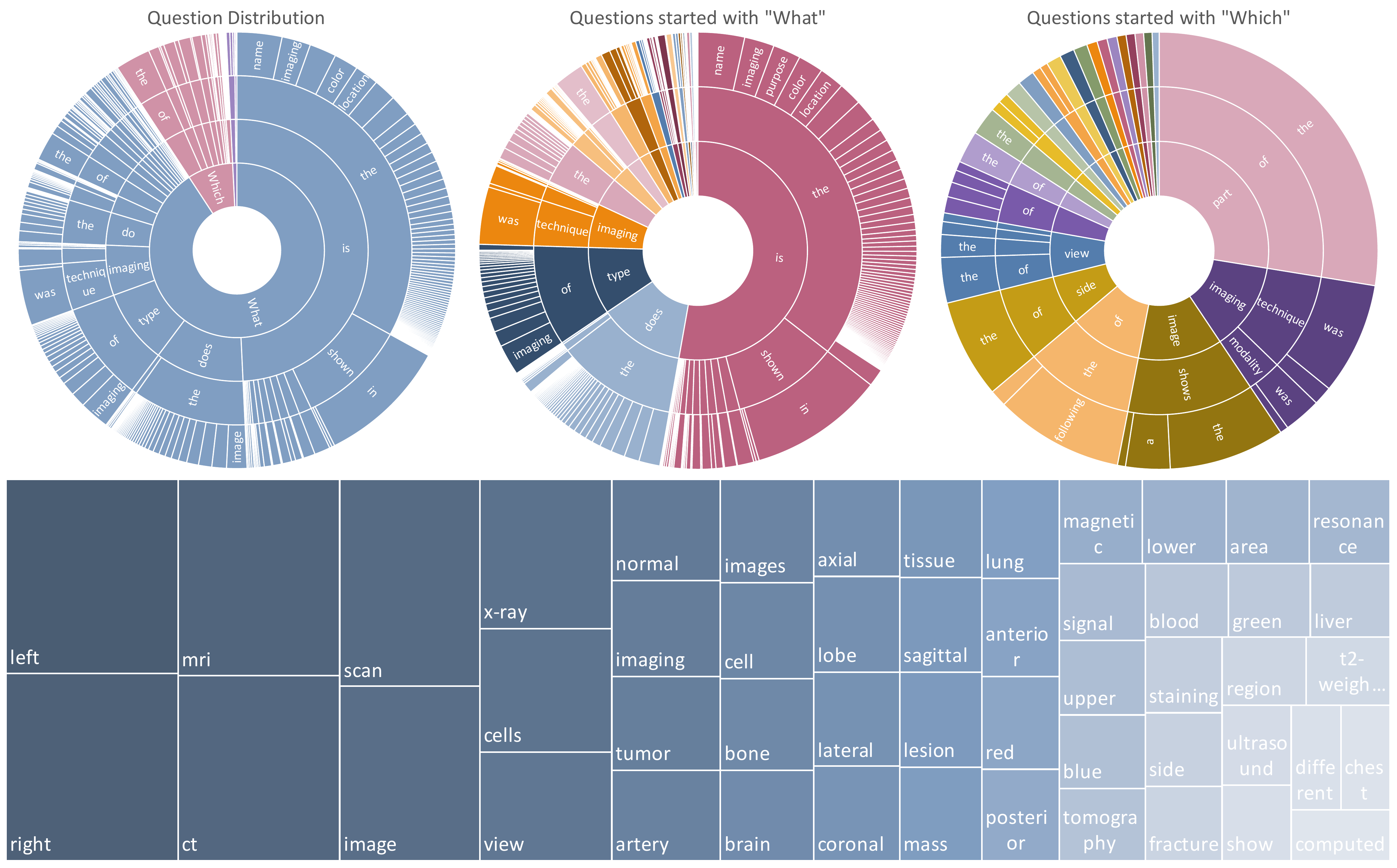}
    \caption{The top row shows the question distribution of the training set by their first four words. From left to right are all questions, questions started with ``What'' and questions started with ``Which''.
    The ordering of the words starts towards the center and radiates outwards.  The bottom row show the answer distribution of the training set. }
    \label{fig:example}
\end{figure}

\subsection{Data Analysis}
This section provides an analysis of images, questions, and answers of our final proposed dataset. 
In detail, the dataset comprises 227k image-question pairs, some examples are presented in Figure~\ref{fig:image_distribution}, which demonstrates the wide diversity of images within our dataset. 
As indicated in Table~\ref{tab:vqa_dataset}, PMC-VQA outperforms existing MedVQA datasets in terms of data size and modality diversity. The questions in our dataset cover a range of difficulties, from simple questions such as identifying image modalities, perspectives, and organs to challenging questions that require specialized knowledge and judgment. Additionally, our dataset includes difficult questions that demand the ability to identify the specific target sub-figure from the compound figure.

Our analysis of the PMC-VQA dataset can be summarized in three aspects: 
(i) \textbf{Images}: We show the top 20 figure types in Figure~\ref{fig:image_distribution}. The images in the PMC-VQA are extremely diverse, ranging from Radiology to Signals. 
(ii) \textbf{Questions}: We clustered the questions into different types based on the words that start the question, as shown in Figure~\ref{fig:example}. The dataset covers very diverse question types, including ``What is the difference...'', ``What type of imaging...'', and ``Which image shows...''. Most questions range from 5 to 15 words, and detailed information about the distribution of question lengths is shown in the supplementary materials~\ref{supple:data_analysis}. 
(iii) \textbf{Answers}: The words in answers primarily encompass positional descriptions, image modalities, and specific anatomical regions. Detailed information about the top 50 words that appeared in the answers is provided in Figure~\ref{fig:example}. Most answers are around 5 words, which is much shorter than the questions. 
The correct options were distributed as follows: A~(24.07$\%$), B~(30.87$\%$), C~(29.09$\%$), D~(15.97 $\%$).

\begin{table*}[t]
\centering
\small
\setlength{\tabcolsep}{5pt}
\renewcommand{\arraystretch}{1.0}
\begin{threeparttable}
\caption{Comparison of existing medical VQA datasets with PMC-VQA, demonstrating our dataset's significant increase in size and diversity. Mixture refers to Radiology, Pathology, Microscopy, Signals, Generic biomedical illustrations, {\em etc}.
}
\label{tab:vqa_dataset}
\vspace{3pt}
\begin{tabular}{lllrr}
\toprule
Dataset  & Modality & Source &  Images &  QA pairs \\
\midrule
VQA-RAD~\cite{lau2018dataset} &  Radiology &  MedPix$^\circledR$ database & 0.3k & 3.5k\\
PathVQA~\cite{he2020pathvqa} &  Pathology & PEIR Digital Library~\cite{jones2001peir} & 5k & 32.8k  \\
SLAKE~\cite{liu2021slake} & Radiology & MSD~\cite{antonelli2022medical}, ChestX-ray8~\cite{wang2017chestx}, CHAOS~\cite{kavur2021chaos} & 0.7k & 14k  \\
VQA-Med-2021~\cite{ben2021overview} & Radiology &  MedPix$^\circledR$ database & 5k & 5k\\
\midrule
PMC-VQA & Mixture\tnote{*} (80\% Radiology) & PubMed Central$^\circledR$ & 149k & 227k \\
\bottomrule
\end{tabular}
\end{threeparttable}
\end{table*}

\subsection{Evaluation on Public Benchmarks}

Table~\ref{tab:transfer} presents the performance of our MedVInT model on three widely recognized MedVQA benchmarks: VQA-RAD, SLAKE, and ImageClef-VQA-2019. 
The results demonstrate that the MedVInT model, regardless of whether we use the ``MedVInT-TE''  or ``MedVInT-TD'' version, surpasses previous best-performing methods on the VQA-RAD and SLAKE datasets. By default, we employ PMC-CLIP as the visual backbone and PMC-LLaMA as the language backbone, as demonstrated in Table~\ref{tab:pmcvqa}, models pre-trained using PubMed Central data generally yield superior performance.

It is important to note that both the VQA-RAD and SLAKE datasets include questions that are categorized as either open-ended or close-ended. Close-ended questions restrict answers to a predefined set of options, whereas open-ended questions allow for free-from text responses. Specifically, for open-ended questions, the accuracy rates were enhanced from 67.2\% to 73.7\% on VQA-RAD and from 81.9\% to 88.2\% on SLAKE. For close-ended questions, the MedVInT model improved the accuracy from 84.0\% to 86.8\%. On the ImageCLEF benchmark, the ``MedVInT-TE'' version of our model achieved a significant improvement with an accuracy rate of 70.5\%, significantly higher than the previous state-of-the-art (SOTA) accuracy of 62.4\%.

\changee{Beyond comparing baselines with their default settings, 
we also consider an architecture-specific comparison where all models are directly trained from scratch on the downstream tasks. 
To distinguish from the default setting, our models here are denoted as ``MedVInT-TE-S'' and ``MedVInT-TD-S''. As shown by the results, our proposed two variants can both surpass the former ``M3AE'' and ``PMC-CLIP'' architectures in most cases.}

Additionally, when comparing the performance of the MedVInT model with and without pre-training on the PMC-VQA-train dataset, using the same architectural framework, it becomes evident that pre-training plays a crucial role in enhancing model performance. Specifically, the ``MedVInT-TE'' version, when pre-trained, showed a remarkable increase of approximately 16\% in accuracy for open-ended questions on VQA-RAD and a 4\% increase on SLAKE, compared to the ``MedVInT-TE-S'' version, which denotes training the model from scratch. Similar enhancements were observed with the ``MedVInT-TD'' version.

\begin{table}[!t]
\centering
\scriptsize
\setlength{\tabcolsep}{1.8pt}
\caption{\changee{Comparison of ACC to SOTA approaches on VQA-RAD, SLAKE, and ImageClef-VQA-2019. 
We use the blank model for evaluation which provides output as free text answers rather than multiple-choice options.
Pre-training data indicates whether the model is pre-trained on the medical multi-modal dataset before training on the target dataset. ``MedVInT-TE-S'' and ``MedVInT-TD-S'' respectively denotes we train the same architecture as ``MedVInT-TE'' or ``MedVInT-TD'' from scratch without pre-training on PMC-VQA.
The best result is bold, the second-best result is underlined.}}
\vspace{3pt}
\begin{threeparttable}
\begin{tabular}{l|c|cccccc}
\toprule
\multirow{2}{*}{Method} &\multirow{2}{*}{{Pretraining Data}} & \multicolumn{2}{c}{VQA-RAD} &  \multicolumn{2}{c}{SLAKE} & VQA-2019 \\  
& & Open  & Close   & Open  & Close   & Overall \\ \midrule
 {M3AE}  & / & 66.5 & 79.0 & 79.2 & 83.4 & - \\
 {PMC-CLIP}& / & 52.0 & 75.4 &  72.7 & 80.0 & - \\
MedVInT-TE-S & /&  53.6~(41.3,64.8) & 76.5~(69.1,84.9)  & 84.0~(80.4,88.4) & 85.1~(79.3,90.1)  & \underline{67.9~(60.6,74.2)}\\
 MedVInT-TD-S & / &  55.3~(45.4,69.4) & 80.5~(74.3,89.4)  & 79.7~(74.6,85.3) & 85.1~(78.2,89.3)  & 58.4~(50.6,66.2) \\
  \midrule
Hanlin & Unknown\tnote{*} & - & - & - &  - & 62.4 \\
 MEVF-BAN & VQA-RAD\tnote{*}~\cite{lau2018dataset} &   49.2 & 77.2  & 77.8 & 79.8 & - \\
 CPRD-BAN & ROCO, MedICaT~\cite{pelka2018roco,subramanian-2020-medicat}&   52.5 & 77.9  & 79.5 & 83.4 &  - \\
 M3AE & CC12M~\cite{changpinyo2021conceptual} &  67.2 & 83.5  & 80.3 & \underline{ 87.8} & - \\
 PMC-CLIP & PMC-OA~\cite{lin2023pmcclip}&  67.0 &  84.0  & 81.9 & {\bf 88.0} & - \\
 \midrule
 MedVInT-TE & PMC-VQA &   \underline{ 69.3~(55.9,79.3)} & \underline{ 84.2~(76.8,90.4)}  & {\bf 88.2~(84.6,92.7)} &  87.7~(81.3,92.8) & {\bf 70.5~(62.8,78.2)}\\
 MedVInT-TD & PMC-VQA & {\bf 73.7~(64.8,84.5)} & {\bf 86.8~(80.4,95.5)} & \underline{ 84.5~(80.4,90.5)} & 86.3~(79.6,90.6) & 61.0~(53.0,67.6)\\
\bottomrule
\end{tabular}
\begin{tablenotes}
\item[*] {``Hanlin'' is a solution in VQA-2019 challenge instead of a detailed scientific paper and, thus, no more details are provided. The numbers are directly copied from challenge papers. ``MEVF-BAN'' views the images in the train set of VQA-RAD as a pretraining dataset, performs image-wise self-supervised learning on it, and finetunes the model with VQA cases on each dataset. . We utilize the results of MEVF-BAN on various VQA benchmarks as reported by PMC-CLIP.}
\end{tablenotes}
\end{threeparttable}
\label{tab:transfer}
\end{table}

\subsection{Evaluation on PMC-VQA}
In this section, we introduce a new MedVQA benchmark, 
termed as PMC-VQA-test. We evaluate different models for both open-ended~(Blanking) and multiple-choice~(Choice) tasks. 
The results are summarized in Table~\ref{tab:pmcvqa}. 
\changeee{GPT-4-Oracle refers to the use of GPT-4 to answer questions based on the original captions of figures in academic papers. This approach represents the upper bound of model performance, as it leverages the most accurate and comprehensive information available about each figure.}
\changee{As shown in the tables, when only using language, 
the model is unable to provide accurate answers and give nearly random outcomes, with an accuracy of only 27.2\% in Blanking and 30.8\% in Choice for LLaMA and enhancing the language model from LLaMA to latest GPT-4 still cannot improve the results, \emph{i.e.}, 21.1\% in Blanking and 25.7\% in Choice for GPT-4. }
\changeee{The lower score in Blanking is due to the language model's tendency to output longer sentences that cannot be correctly matched to a specific choice, which affects the calculation of model's accuracy.}
It is worth noting that around 30\% of the questions have ``B'' answers, making the 30.8\% score nearly equivalent to the highest possible score attainable through guessing. These observations highlight the crucial requirement of multimodal understanding in our dataset and emphasize the strong relationship between images and the questions posed. In contrast to the training split, PMC-VQA-test has undergone thorough manual checking~(Check { Sec.~\ref{sec:pmcvqa}} for more details), 
ensuring the credibility of the evaluation. 
We also report the experimental results on the original randomly split test set \changeee{PMC-VQA-test-initial}, which is larger but lacks further manual checking, in the supplementary materials~\ref{supple:eval_on_original}.

\begin{table}[t]
\centering
\scriptsize 
\setlength{\tabcolsep}{3pt}
\caption{
\changeee{
Comparison of baseline models using different pre-trained models on both open-ended (Blank) and multiple-choice (Choice) tasks. We reported the results of the PMC-VQA-test. ``Scratch'' means to train the vision model from scratch with the same architecture as PMC-CLIP. }
}
\renewcommand\arraystretch{1.05}
\begin{tabular}{lllccc}
\toprule
\multirow{2}{*}{\textbf{Method}}                                    & \multirow{2}{*}{\textbf{Language Backbone}}          & \multirow{2}{*}{\textbf{Vision Backbone}}                      & Choice      & \multicolumn{2}{c}{Blanking}                   \\ \cmidrule{4-6} 
                                                                    &                                                      &                                                       & ACC   &   ACC & BLEU-1 \\  \midrule
\multicolumn{6}{l}{\textbf{Language-only}}\\ \midrule
{GPT-4-Oracle~\cite{openai2023gpt4}}                    &  {GPT-4~\cite{openai2023gpt4}}         &        {--}      &  89.3 (87.7,90.8) & 22.0 (19.6,24.5) & 18.8 (17.6,20.2)  \\
{GPT-4~\cite{openai2023gpt4}}                    &  {GPT-4~\cite{openai2023gpt4}}         &        {--}      &   {25.7 (23.5,28.1)}       &       {21.1(18.8,23.5)}       &      {3.0(2.6,3.4)}    \\ 
LLaMA~\cite{touvron2023llama}                    & LLaMA~\cite{touvron2023llama}     & --                & {30.8 (27.4,34.8)}         & {27.2 (23.1,31.3)}              & {14.6 (12.7,16.6)}          \\  \midrule
\multicolumn{6}{l}{\textbf{Zero-shot}} \\ \midrule
PMC-CLIP~\cite{lin2023pmcclip}                   & PMC-CLIP~\cite{lin2023pmcclip}    & PMC-CLIP~\cite{lin2023pmcclip}     &  24.7 (21.3,28.0)      & - & -     \\  
BLIP-2~\cite{li2023blip}                               & OPT-2.7B~\cite{Zhang2022OPTOP}    & CLIP~\cite{radford2021learning}    &  24.3 (20.7,27.7)      & {21.8 (17.2,26.4)}              & {7.6 (5.3,9.9)}               \\  
Open-Flamingo~\cite{awadalla2023openflamingo}    & LLaMA~\cite{touvron2023llama}           & CLIP~\cite{radford2021learning}    &  26.4 (22.7,29.8)        & {26.5 (22.3,30.7)}              & {4.1 (2.1,6.13)}        \\ 
{LLaVA-Med~\cite{li2024llava}} & {Vicuna~\cite{vicuna2023}} & {BioMedCLIP~\cite{zhang2023biomedclip}} & {34.8 (32.2,37.8)} & {29.4 (26.6,32.1)} & {3.9(3.5,4.2)}  \\
MedICap-GPT-4 & {GPT-4~\cite{openai2023gpt4}}  & MedICap~\cite{nicolson2023concise} & 27.2 (24.7,29.7) & 20.9 (18.8,23.3) & 4.2 (3.6,4.6) \\
\midrule
\multicolumn{6}{l}{\textbf{Trained on PMC-VQA}}\\ \midrule
MedICap-PMCVQA-GPT-4 & {GPT-4~\cite{openai2023gpt4}}  & MedICap-PMCVQA & 35.9 (33.0, 38.3) & 22.4 (20.1,24.8) & 3.8 (3.3,4.3) \\
\midrule
\multirow{9}{*}{MedVInT-TE}                                         & \multirow{3}{*}{PubMedBERT~\cite{gu2021domain}}        & Scratch                                               &  34.9 (31.7,38.5)  & {34.2 (31.2,37.0)}  & {20.9 (18.9,23,2)} \\  
                                                                    &                                                      & CLIP~\cite{radford2021learning}     & 34.3 (30.7,37.8) & {34.4 (31.0,37.6)}  & {20.8 (18.6,23.3)} \\  
                                                                    &                                                      & PMC-CLIP~\cite{lin2023pmcclip}     & 37.6 (34.7,40.9) & \underline{36.4 (32.6,39.4)}  & \textbf{23.2 (21.2,25.7)} \\   \cmidrule{2-6} 
                                                                    & \multirow{3}{*}{LLaMA-ENC~\cite{touvron2023llama}}     & Scratch                                                      & 35.2 (31.8,38.3)  & {32.5 (29.6,35.9)}  & {15.9 (12.8,16.8)} \\  
                                                                    &                                                      & CLIP~\cite{radford2021learning}         & 36.1 (31.0,39.5) & {33.4 (29.8, 36.5)} & {15.1 (12.8,17.5)} \\
                                                                    &                                                      & PMC-CLIP~\cite{lin2023pmcclip}        & 37.1 (34.0,40.1) & \textbf{36.8 (33.5,40.0)}  & {18.4 (15.6,20.5)}   \\ \cmidrule{2-6} 
                                                                    & \multirow{3}{*}{PMC-LLaMA-ENC~\cite{wu2023pmcllama}}   & Scratch                                           & 38.0 (34.9,42.2) & {35.0 (31.9,38.5)}  & {17.0 (14.5,18.9)} \\    
                                                                    &                                                      & CLIP~\cite{radford2021learning}            & 38.5 (35.7,42.4) & {34.4 (31.3,37.8)}  & {16.5 (14.4,18.8)} \\   
                                                                    &                                                      & PMC-CLIP~\cite{lin2023pmcclip}          & 39.2 (36.7,41.7)    & {35.3 (31.4, 38.8)} & {18.6 (16.6,21.6)}        \\  \midrule
\multirow{6}{*}{MedVInT-TD}                                         & \multirow{3}{*}{LLaMA~\cite{touvron2023llama}}               & Scratch                                                       & 37.9 (34.5,41.4)    &      {30.2 (26.9,33.8)}              & {18.0 (16.2,20.0)}             \\  
                                                                    &                                                      & CLIP~\cite{radford2021learning}         & 39.2 (35.3,42.7)       & {32.2 (29.4,36.0)}              & {20.0 (17.8,23.0)}             \\   
                                                                    &                                                      & PMC-CLIP~\cite{lin2023pmcclip}            & \underline{39.5 (35.1,42.7)}     & {33.4 (30.6,37.4)}              & {21.3 (18.9,23.8)}                \\ \cmidrule{2-6} 
                                                                    & \multirow{3}{*}{PMC-LLaMA~\cite{wu2023pmcllama}}      & Scratch                                                       & 36.9 (33.2,40.2)       & {29.8 (26.9,32.7)}              & {17.4 (15.1,19.6)}                   \\  
                                                                    &                                                      & CLIP~\cite{radford2021learning}            & 36.9 (32.9,40.1)    & {32.6 (29.0,36.2)}              & {20.4 (18.1,22.9)}                    \\ 
                                                                    &                                                      & PMC-CLIP~\cite{lin2023pmcclip}             & \textbf{40.3 (37.2,43.8)}     & {33.6 (29.9,36.5)}              & \underline{21.5 (19.4,24.0)}                       \\   \bottomrule
\end{tabular}
\label{tab:pmcvqa}
\end{table}

\begin{figure}[!tbh]
    \centering
    \includegraphics[width=\textwidth]{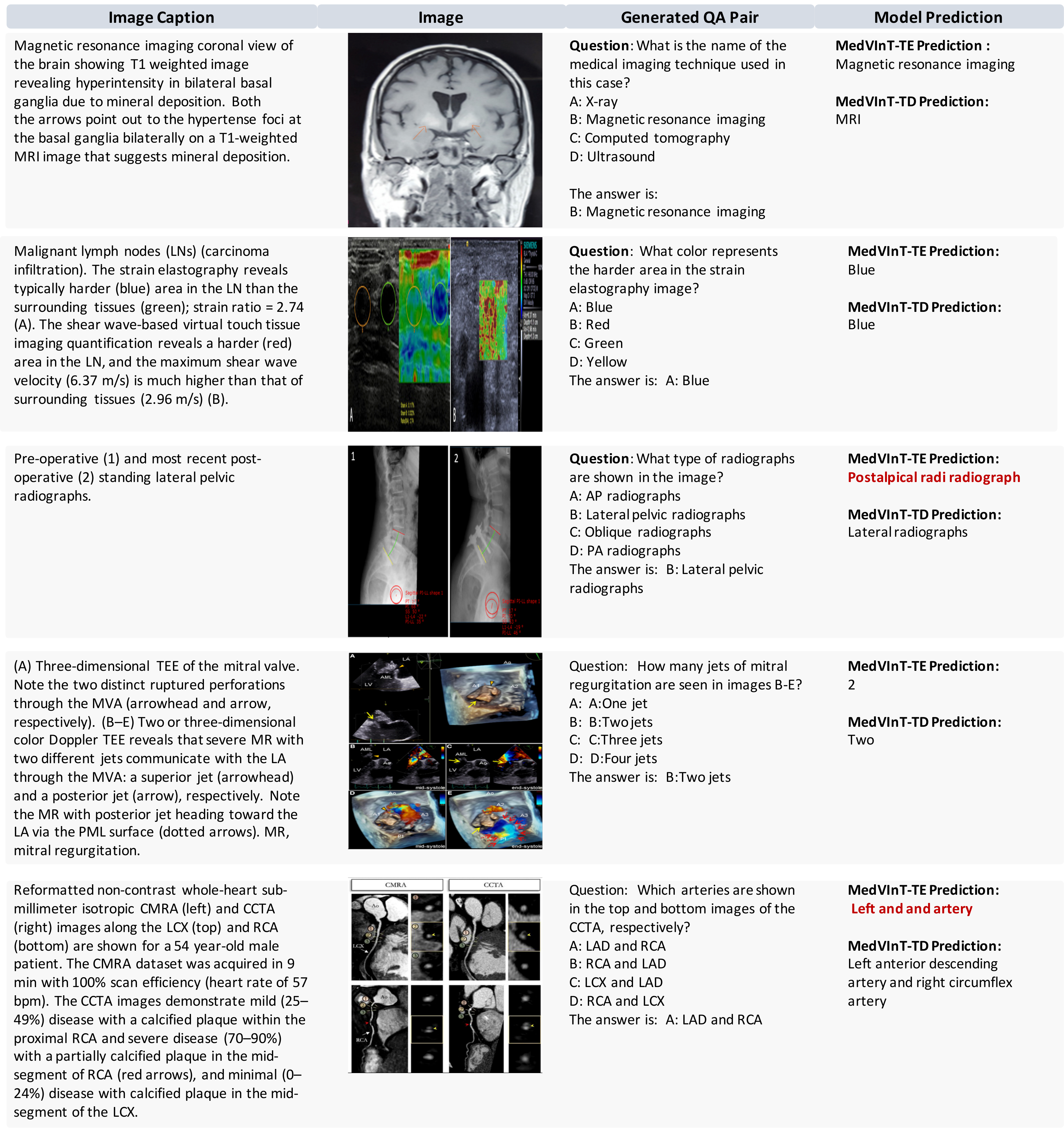}
    \caption{Examples of image captions, images, the generated question-answer pairs, and model prediction. The wrong predictions are highlighted in red.}
    \label{fig:sample}
\end{figure}

\changee{We also present the zero-shot evaluation results of the general VQA models like PMC-CLIP, BLIP-2, and Open-Flamingo which show relatively lower performance on the choice task. 
For instance, in the choice task, the model Open-Flamingo only achieved a 26.4\% accuracy rate, significantly lower performance than our model at 40.3\%.  We also evaluate the medical-specific generative-based VQA model, {\em e.g.}, LLaVA-Med. Though it is better than the general models, it still lags behind our proposed MedVinT. It's worth noting that LLaVA-Med is a work after our first announcement.}
This contrasts with the trained models on PMC-VQA, where we see notable improvements. Specifically, the MedVInT-TE and MedVInT-TD models, when paired with the PMC-CLIP vision backbone, demonstrate superior performance. 
For the open-ended task, the PMC-CLIP vision backbone again proves beneficial, with the MedVInT-TE model reaching the highest accuracy (36.4\%) and BLEU-1 score (23.2\%) when combined with the PubMedBERT language backbone. Moreover, the comparison between models trained from scratch and those utilizing CLIP or PMC-CLIP as vision backbones across different configurations of language backbones (PubMedBERT, LLaMA-ENC, and PMC-LLaMA-ENC) reveals a consistent trend: pre-trained models, especially those pre-trained with domain-specific data (PMC-CLIP), tend to outperform their counterparts trained from scratch. This emphasizes the importance of pre-training in achieving higher accuracies and better natural language generation metrics in MedVQA tasks. We then prompted a Large Language Model (LLM) to answer questions based on these generated captions. 
\changeee{We also compared our approach with two-stage Visual Question Answering (VQA) models, which employ image captioning followed by a large language model for question answering. We experimented with a two-stage VQA method similar to  Chatcad~\cite{wang2023chatcad}. We first used MedICap~\cite{nicolson2023concise}, a state-of-the-art medical image captioning model, to interpret the given images into captions.
The results showed poor performance on the test set. 
We then trained MedICap on the original image-caption pairs from the PMC-VQA training set to mitigate the domain gap. 
As shown, MedICap-PMCVQA-GPT-4 still shows inferior performance, which highlights key challenges in the two-stage approach: Captioning models need to anticipate potential questions in their descriptions. There's often a mismatch between caption content and question focus. For example, a caption might state ``This is an MRI image of a brain.'' while the question asks ``Is there a mass in the image?''.
}
To provide a more comprehensive understanding of the dataset, we offer additional examples illustrated in Figure~\ref{fig:sample}. This figure showcases random instances of the original image and corresponding captions, along with multiple-choice questions generated from them.

\changeee{\subsection{Evaluation of Visual Backbone Performance}
We conducted additional experiments on standard medical image classification tasks to demonstrate the visual backbone's performance and its improvement through the VQA pre-training. 
We evaluated our model on the MedMNIST dataset \cite{yang2021medmnist}, which provides a diverse set of medical imaging modalities and classification tasks. 
}

\begin{table*}[htpb]
    \centering
    \scriptsize 
    \setlength{\tabcolsep}{15pt}
    \caption{\changeee{Classification results on three representative subsets of MedMNIST: PneumoniaMNIST (chest X-ray), BreastMNIST (ultrasound), and DermaMNIST (dermatoscopy). The best results are in \textbf{bold}, and the second-best are in \underline{underlined}.}}
    \vspace{3pt}
    \begin{tabular}{l|cc|cc|cc}
    \toprule
        \multirow{2}{*}{Methods} & \multicolumn{2}{c|}{PneumoniaMNIST} & \multicolumn{2}{c|}{BreastMNIST} & \multicolumn{2}{c}{DermaMNIST} \\
         & AUC$\uparrow$ & ACC$\uparrow$ & AUC$\uparrow$ & ACC$\uparrow$ & AUC$\uparrow$ & ACC$\uparrow$ \\
        \midrule
        ResNet50 \cite{he2016deep} & 96.20 & 88.40 & 86.60 & 84.20 & 91.20 & 73.10 \\
        DWT-CV \cite{cheng2022dwt} & 95.69 & 88.67 & 89.77 & 85.68 & 91.67 & 74.75 \\ 
        SADAE \cite{ge2022self} &  98.30 &  91.80 &  91.50 &  87.80 &  92.70 &  75.90 \\
        PMC-CLIP &  \textbf{99.02} &  \textbf{95.35} &  \textbf{94.56} &  \textbf{91.35} &  \underline{93.41} &  \underline{79.80} \\
        \midrule
        MedVInT-TE & \underline{98.49} & \underline{94.87} & \underline{93.44} & \underline{90.38} & \textbf{93.71} & \textbf{80.00} \\ 
        MedVInT-TD & 97.39 & 94.71 & 90.04 & 87.82 & 93.43 & 78.30 \\
     \bottomrule
    \end{tabular}
    \label{tab:MedMNIST}
\end{table*}

\changeee{As shown in Table \ref{tab:MedMNIST}, our MedVInT models demonstrate competitive performance across all three tasks. 
Notably, MedVInT-TE achieves the best performance on DermaMNIST and the second-best performance on PneumoniaMNIST and BreastMNIST, only slightly behind PMC-CLIP. The results are impressive considering that MedVInT was pre-trained on only 177K images, compared to PMC-CLIP's 1.6M image-caption pairs.
Our results demonstrate the effectiveness of VQA-based pre-training compared to CLIP-style training. While both approaches aim to align visual and textual information, VQA requires a deeper understanding of the image content to answer specific questions.
This difference in training objectives appears to lead to more robust visual representations, as evidenced by our model's competitive performance despite being trained on significantly fewer images.
These results demonstrate that our MedVQA task not only ``standardizes'' data into QA pairs but also substantially improves the visual backbone's performance on various medical image classification tasks. 
}
\section{Discussion}

In this study, we target the challenge of MedVQA, where even the strongest VQA models trained on natural images yield results that closely resemble random guesses. To overcome this, we propose MedVInT, 
a generative model tailored to advance this crucial medical task.
MedVInT is trained by aligning visual data from a pre-trained vision encoder with language models. 
Additionally, we present a scalable pipeline for constructing PMC-VQA, a comprehensive VQA dataset in the medical domain comprising 227k pairs across 149k images, spanning diverse modalities and diseases. 
Our proposed model delivers state-of-the-art performance on existing datasets, providing a new and reliable benchmark for evaluating different methods in this field. 

\changeee{\noindent \textbf{Significance of Medical VQA for Medical Imaging Ecosystem.}
The development of advanced MedVQA systems has far-reaching implications for various stakeholders in the medical imaging ecosystem~\cite{bajwa2021artificial,yang2023impact,clusmann2023future}. 
For radiologists and referring physicians, MedVQA can serve as a powerful decision-support tool, potentially enhancing diagnostic precision and streamlining image interpretation processes~\cite{demirhan2023survey}. This could lead to more efficient clinical workflows and allow healthcare professionals to dedicate more time to direct patient care.
For patients, MedVQA systems can significantly improve the communication of complex medical information. By translating intricate radiology reports into more comprehensible language, these systems can enhance patient understanding and engagement in their healthcare journey. This aligns with the growing emphasis on patient-centered care and shared decision-making in modern healthcare practices~\cite{park2024patient}.
From a research and education perspective, MedVQA systems like MedVInT, trained on comprehensive datasets such as PMC-VQA, can serve as valuable tools for medical students and researchers~\cite{safranek2023role}. They can provide interactive learning experiences, assist in the design of research plans, and offer insights into complex medical imaging concepts, thereby contributing to the advancement of medical knowledge and skills.}

\noindent \textbf{PMC-VQA Act as a Valuable Resource for Medical VQA Domain.}
Previous MedVQA datasets are usually limited in size and diversity, as demonstrated in Table~\ref{tab:vqa_dataset}.
In contrast, PMC-VQA represents a pivotal advancement, offering an extensive resource that addresses the diverse and complex needs of the medical VQA domain. 
PMC-VQA facilitates the development of models capable of understanding and interpreting medical imagery with unprecedented accuracy and detail.
Moreover, \changee{comparing} results using the same architecture, with and without PMC-VQA (Table~\ref{tab:pmcvqa}),  it is clear that pre-training with PMC-VQA significantly outperforms. 
These results highlight the critical role that our PMC-VQA plays in addressing the major challenges that hinder the development of a generative MedVQA system.
The pre-training enables models to gain a deep understanding of medical visuals and their associated questions, significantly enhancing their predictive capabilities.

\noindent \textbf{General Visual-language Models Struggle on MedVQA.}
We evaluated the zero-shot performance of existing SOTA multimodal models, 
BLIP-2 and open-source version of Flamingo~\cite{li2023blip,awadalla2023openflamingo}.
As shown, even the best-performing models in natural images struggle to answer our questions, demonstrating the challenging nature of our dataset and its strong biomedical relevance. 
These results highlight the critical role that our PMC-VQA-train plays in addressing the major challenges that hinder the development of a generative MedVQA system.

\noindent \textbf{MedVInT Achieves State-of-the-art Performance of Generative MedVQA.}
As demonstrated in the results, both MedVInT-TE and MedVInT-TD perform well on the MedVQA tasks. 
We compared it against various baselines that use different generative model backbones. 
Our results show that replacing the general visual backbone with a specialized medical one leads to improved performance, highlighting the importance of visual understanding in our test set. Additionally, we observed that replacing the language backbone with a domain-specific model also leads to some improvements, although not as significant as those achieved in the visual domain.
In addition, the gap between the two training styles mainly exists in open-ended questions, with ``MedVInT-TD'' performing better on VQA-RAD and ``MedVInT-TE'' being more effective on SLAKE. 
This difference can be attributed to the fact that the VQA-RAD answers are typically longer than those in SLAKE, making the ``MedVInT-TD'' model more suitable. 
Conversely, SLAKE questions often require short responses, making the ``MedVInT-TE'' model more appropriate for such retrieve-like tasks.

\noindent \textbf{PMC-VQA-test Presents a Significantly More Challenging Benchmark.}
Notably, the previous SOTA medical multimodal model, 
PMC-CLIP~\cite{lin2023pmcclip}, struggles on our dataset. 
Not only does it fail to solve the blanking task, but it also significantly underperforms on multi-choice questions, with accuracy close to random. 
These findings underline the difficulty of our proposed benchmark and its capacity to provide a more rigorous evaluation of VQA models. 
However, while evaluating our proposed challenging benchmark, 
even the state-of-the-art models struggle, showing that there is still ample room for development in this field.



\noindent \textbf{Impacts of Our Work.}
Since released to the public, we are delighted to observe the rapid adoption and extensive utilization of the PMC-VQA dataset, across a diverse range of research endeavors since its release. 
The dataset has served as a foundational resource for the development of numerous generative models, demonstrating its significant impact on the field. Notable examples include MathVista~\cite{lu2023mathvista}, RadFM~\cite{wu2023towards}, 
Qilin-Med-VL~\cite{liu2023qilin}, SILKIE~\cite{li2023silkie}, CheXagent~\cite{chen2024chexagent}, UniDCP~\cite{zhan2023unidcp}, and Quilt-LLaVA~\cite{seyfioglu2023quilt}. 
In addition, the methodology employed in constructing the dataset and the innovative prompt strategies we introduced have also inspired a series of works~\cite{wu2024hallucination} and ~\cite{chen4578568chatffa}.
Furthermore, many studies have compared with our proposed MedVInT, recognizing it as the pioneering medical generative foundation model, such as Med-flamingo~\cite{moor2023med}, OmniMedVQA~\cite{hu2024omnimedvqa}.
This widespread adoption not only validates the robustness and utility of our dataset but also highlights its role in the scientific community.

\noindent \textbf{Limitations.}
The proposed PMC-VQA, while comprehensive, is subject to several limitations. First, similar to all existing datasets, there might be potential distribution biases in the images included in PMC-VQA compared to clinical practice. Specifically, our data is curated from academic papers, where there may be selective use of images to illustrate typical cases or slices, along with additional annotations such as arrows to aid understanding, resulting in our data being simpler compared to clinical scenarios.
Nevertheless, for training purposes, the data from PMC-VQA remains crucial to help models better understand real clinical imaging data, as shown by the performance on public benchmarks in Table~\ref{tab:transfer}. On the other hand, for testing, \emph{i.e.}, the benchmark we propose as shown in Table~\ref{tab:pmcvqa}, even in such relatively simple scenarios, current methods still face significant challenges. Hence, for the ongoing advancement of MedVQA, conducting assessments in such an experimental playground to steer the emergence of more potent methodologies for the future still holds significance. On evaluation metrics, measuring the results from generative models poses a general challenge in the entire AI community~\cite{vicuna2023}, and this holds true for our testing as well. Although both the ACC score and Bleu score are used in our benchmark for assessing open-ended blanking results, these two metrics fail to capture the fluency of the generated sentence since they measure string similarity irrespective of word order. 
The encoder-based model thus significantly underperforms the decoder-based model in this regard. To address this issue, 
we plan to explore more accurate and effective evaluation metrics in our benchmark in future work.
\changee{
Lastly, as a starting point for generative-based  MedVQA methods, our models may still suffer from hallucinations in non-sensical or adversarial cases with huge domain gaps~(more case studies in our supplementary). Thus this paper is more as a \textbf{proof-of-concept} for building generative-based medical VQA models and needs more future efforts for real clinical applications. 
}

\section{Method}
\subsection{The PMC-VQA Dataset}
\label{sec:pmcvqa}
\begin{figure}[!htb]
    \centering
    \includegraphics[width=0.98\textwidth]{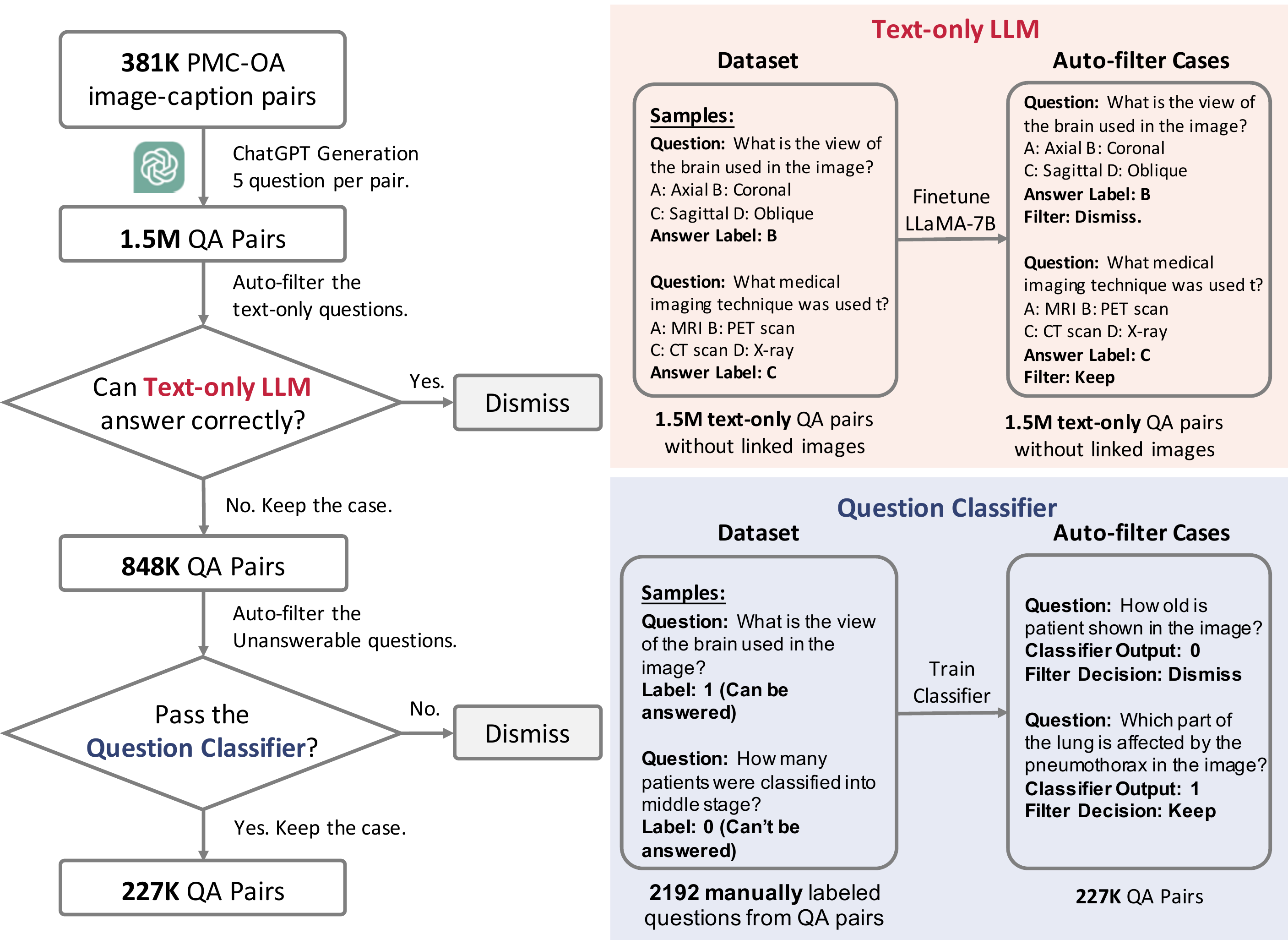}
    \vspace{3pt}
    \caption{The whole flowchart demonstrating how we build up our PMC-VQA dataset. In left, we show the general progress and in right we show how we build up the two auto-filter models used in our data collection. }
    \label{fig:workflow}
\end{figure}

Our study has identified the lack of large-scale, multi-modal MedVQA datasets as a significant obstacle to the development of effective generative MedVQA models. 
In this section, we provide a detailed description of our dataset collection process, starting with the source data and continuing with the question-answer generation and data filtering procedures. 
Finally, we analyze the collected data from various perspectives to gain insights into its properties and potential applications. The main data collection flow can be found in Figure~\ref{fig:workflow}.

\noindent {\bf Source Data. }
We start from PMC-OA~\cite{lin2023pmcclip}, which is a comprehensive biomedical dataset comprising 1.6 million image-text pairs collected from PubMedCentral~(PMC)'s OpenAccess subset~\cite{roberts2001pubmed}, covering 2.4 million papers. 
The pipeline of creating PMC-OA consists of three major stages: {i} medical figure-caption collection; (ii) subfigure separation; (iii) subcaption separation \& alignment. To maintain the diversity and complexity of PMC-VQA, we have used a version of \textbf{381K image-caption pairs} obtained from the first stage of the medical figure collection process without subfigure auto-separation.


\noindent {\bf Question-Answer Generation. }
To automatically generate high-quality question-answer pairs, we input the image captions of PMC-OA, and prompt ChatGPT to generate 5 question-answer pairs for each caption. 
We use the following prompt to generate 5 question-answer pairs for each caption.

\colorbox[RGB]{230,230,230}{
    \parbox{0.95\textwidth}
    {
    Ask 5 questions about the content and generate four options for each question. The questions should be answerable with the information provided in the caption, and the four options should include one correct and three incorrect options, with the position of the correct option randomized. The output should use the following template: i:`the question index' question:`the generate question' choice: `A:option content B:option content C:option content D:option content'  answer: The correct option(A$\backslash$B$\backslash$C$\backslash$D).  
    }
}

This approach allows us to generate a large volume of diverse and high-quality questions that cover a wide range of medical topics.
Considering some captions are too short to ask 5 questions, ChatGPT will repeat generated question-answer pairs or refuse to generate new pairs halfway and we dismissed the dummy cases.
After generating the question-answer pairs using ChatGPT, we applied a rigorous filtering process to ensure that the pairs met our formatting requirements. 
As a result, we obtained 1,497,808 question-answer pairs, and since the original captions are linked with images, the pairs can naturally find corresponding images, resulting in an average of 3.93 pairs per image.

\noindent {\bf Automatic \& Manual Data Filtering. }
As the questions are sourced from image captions,
some of them can be answered correctly using biomedical knowledge alone, 
{\em i.e.}, without the need for a specific image, for example, 
question: ``which type of MRI sequence shows high signal in the marrow edema?''.
To address this issue, we trained a question-answer model using LLaMA-7B~\cite{touvron2023llama} with \textbf{text data only} and eliminated all questions that could be potentially answered by the language model. 
Specifically, we first split the dataset into two parts, then we train a LLaMA-7B model only text input following the full fine-tuning pipeline introduced in PMC-LLaMA~\cite{wu2023pmcllama} in each part and do inference on the other part. To avoid that sometimes language model may make the correct choice by randomly guessing, for each case, we will shuffle the choice list and do inference five times. The questions the language model can make the right choice three times out of five will be dismissed.
This filtering process resulted in 848,433 question-answer pairs that are unanswerable by the language-only model.

Furthermore, some questions in our data rely on additional information in the caption that cannot be answered with only the corresponding image, such as ``How many patients were classified into the middle stage?" 
To identify these questions, we manually annotated 2192 question-answer pairs with binary labels, using `1' for answerable based on images and `0' otherwise. Then we train and evaluate a question classification model on these labeled data, specifically 1752 pairs for training and 440 for testing, and the model can achieve an accuracy of 81.77\% on this binary classification task. 
We then used this model for data cleaning, resulting in a total of 226,946 question-answer pairs corresponding to 149,075 images, termed as {\bf PMC-VQA} dataset.

From this cleaned dataset, we randomly selected 50,000 image-question pairs to create an initial test set, \changeee{PMC-VQA-test-initial}.
The same image is guaranteed to not appear in both the training and testing sets.
Additionally, we manually checked some test samples again, resulting in a small clean test set of 2,000 samples, which were \textbf{manually} verified for quality, termed as {\bf PMC-VQA-test}, where we mainly consider the following criteria:
\vspace{-0.2cm}
\begin{itemize}
\setlength\itemsep{0.15cm}
    \item whether questions are related to the image and can be answered via images;
    \item whether the distractor choices in the candidate list are complex enough, to avoid pure guessing from options;
    \item whether the image quality is good enough, dismissing the ``paper images'' which contain too many extra elements~(charts, flows or numbers).
\end{itemize}
During this verification procedure, we have estimated that over 80\% cases in {\bf PMC-VQA-test} can be retained.


\subsection{Architecture Design}
\begin{figure}[t]
    \centering
    \includegraphics[width=\textwidth]{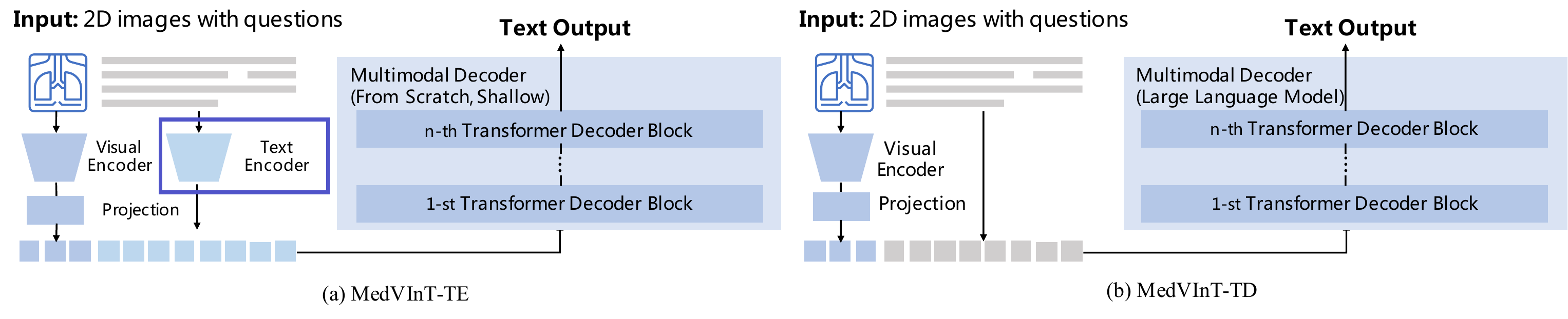}
    \caption{\change{The proposed architecture, mainly consists of three components: a visual encoder to extract visual features, a text encoder to encode textual context, and a multimodal decoder to generate the answer. (a) MedVInT-TE, encodes textual context~(blue box) before input to the multimodal decoder; (b) MedVInT-TD, concatenates text tokens with visual features as input.}}
    \label{fig:train_styles}
\end{figure}

We start with an introduction to the problem of generative medical visual question answering in Sec.~\ref{sec:problem}, and detail our proposed architecture for generative MedVQA~(Figure~\ref{fig:train_styles}). We mainly focus on leveraging the pre-trained uni-model model to build up a multi-modal generative VQA achitecture. Specifically, we offer two model variants,  that are tailored to encoder-based and decoder-based language models, respectively, denoted as MedVInT-TE~(Sec.~\ref{sec:MedVInT-TE}) and MedVInT-TD~(Sec.~\ref{sec:MedVInT-TD}).

\subsubsection{Problem Formulation}
\label{sec:problem}
MedVQA is a task of answering natural language questions about medical visual content, 
typically images or videos obtained from medical devices like X-ray, CT, MRI, or microscopy, {\em etc.}
Specifically, our goal is to train a model that can output the corresponding answer for a given question, which can be expressed as:
\begin{equation}
    \hat{a}_i = \Phi_{\text{MedVQA}}(\mathcal{I}_i, q_i; \Theta) = \Phi_{\text{dec}}(\Phi_{\text{vis}}(\mathcal{I}_i; \theta_{\text{vis}}), \Phi_{\text{text}}(q_i; \theta_{\text{text}}); \theta_{\text{dec}}) 
\end{equation}
Here, $\hat{a}_i$ refers to the predicted answer,
$\mathcal{I}_i \in \mathbb{R}^{H \times W \times C}$ refers to the visual image, $H,W,C$ are height, width, channel respectively. 
The posed question and corresponding ground-truth answer in the form of natural language are denoted as $q_i$ and $a_i$, respectively. $\Theta = \{\theta_{\text{vis}}, \theta_{\text{text}}, \theta_{\text{dec}}\}$ denote the trainable parameters. 

Existing approaches have primarily treated medical VQA as a classification problem, with the goal of selecting the correct answer from a candidate set, {\em i.e.}, $a_i \in \Omega = \{a_1, a_2, \dots, a_N\}$, where $N$ represents the total number of answers within the dataset. Consequently, 
this approach limits the system's utility to predefined outcomes, hampering its free-form user-machine interaction potential.

In this paper, we take an alternative approach, 
with the goal of generating an open-ended answer in natural language.
Specifically, we train the system by maximizing the probability of generating the ground-truth answer given the input image and question. The loss function used to train the model is typically the negative log-likelihood of correctly inferring the next token in the sequence, summed over all \change{token} steps, expressed as: 
\begin{equation}
\mathcal{L}(\Theta) = -\sum_{t=1}^{T} \log p(a^t|\mathcal{I}, q^{1:T}, a^{1:t-1}; \Theta)
\end{equation}
where $T$ is the length of the ground-truth answer, and $p(a^t|\mathcal{I}, q^{1:T}, a^{1:t-1}; \Theta)$ is the probability of generating the $t$-th token in the answer sequence given the input image $\mathcal{I}$, the question sequence $q^{1:T}$, and the previous tokens in the answer sequence $a^{1:t-1}$. 
This formulation allows the model to generate diverse and informative answers, which can be useful in a wider range of scenarios than traditional classification-based methods.

\subsubsection{ MedVInT-TE.}
\label{sec:MedVInT-TE}
\textbf{Visual Encoder.} 
Given one specific image $\mathcal{I}$, 
we can obtain the image embedding, {\em i.e.}, $\boldsymbol{v} = \mathrm{\Phi}_\text{vis}(\mathcal{I}) \in \mathbb{R}^{n\times d}$,
where $d$ denotes the embedding dimension, $n$ denotes the patch number.
The vision encoder is based on a pre-trained ResNet-50 adopted from PMC-CLIP~\cite{lin2023pmcclip}, with a trainable projection module.
We propose two distinct variants for this projection module. 
The first variant, MLP-based, employs a two-layer Multilayer Perceptron (MLP), while the second variant, transformer-based, employs a 12-layer transformer decoder supplemented with several learnable vectors as query input.

\textbf{Text Encoder.} 
Given one question on the image, we append a fixed prompt with the question to guide the language model with desirable output, {\em i.e.}, ``Question: $\{question \}$, the answer is: '', and encode it with the language encoder: $\boldsymbol{q} = \mathrm{\Phi}_\text{text}(q) \in \mathbb{R}^{l\times d}$,
where $\boldsymbol{q}$ refers to the text embedding, 
$l$ represents the sequence length for the prompt, 
and $q$ is the prompted question.
$\Phi_\text{text}$ is initialized with the pre-trained language model.
Note that our model can also be applied to multiple-choice tasks, 
by providing options and training it to output the right choice as "A/B/C/D". 
The prompt is then modified as ``Question: $q$, the options are: $a_1, a_2, a_3, a_4$, the answer is: '', where $a_i$ refers to the $i$-th option.


\textbf{Multimodal Decoder.}
With encoded visual embeddings~($\boldsymbol{v}$) and question embeddings~($\boldsymbol{q}$), we concatenate them as the input to the multimodal decoder~($\mathrm{\Phi}_\text{dec}$). The multimodal decoder is initialized from scratch with a 4-layer transformer structure. Additionally, acknowledging that the encoder-based language models lack casual masking, we reformulate the generation task as a mask language modeling task, {\em i.e.}, the question input is padded with several `[MASK]' token and the decoder module learns to generate the prediction for the masked token.

\subsubsection{ MedVInT-TD.}
\label{sec:MedVInT-TD}

\noindent  \textbf{Visual Encoder.} 
The visual encoder is the same as in MedVInT-TE.

\textbf{Text Encoder.} 
We design $\Phi_\text{text}$ as a simple tokenization embedding layer, 
similar to the primary GPT-like LLMs, and the tokenization layer can be initialized with the corresponding layer of any chosen pre-trained LLM, like LLaMA~\cite{touvron2023llama} or PMC-LLaMA~\cite{wu2023pmcllama}. Same with MedVInT-TE, it also encodes the question input into embedding features $\boldsymbol{q}$ and can perform multi-choice or blank through different prompts.

\textbf{Multimodal Decoder.}
For the Transformer decoder-based language model, with its output format already being free-form text, we directly use its architecture as the multimodal decoder initialized with the pre-trained weights. 
Specifically, we concatenate the image and text features as the input.
However, directly using the text decoder as a multimodal decoder, may lead to significant mismatching between the image encoding space and the decoder input space. Therefore, to further fill the gap between the image embedding space, here, we pre-train the whole network with the PMC-OA~\cite{lin2023pmcclip} dataset by captioning each image, which is similar to BLIP-2~\cite{li2023blip}. \change{Then train for the MedVQA task on our PMC-VQA dataset.}

\subsection{Datasets and Backbones}
\subsubsection{Existing MedVQA Datasets}
\label{sec:dataset}
In the paper, we evaluate our final model MedVInT on three main public benchmarks, namely VQA-RAD, SLAKE, and ImageClef-VQA-2019.

\noindent \textbf{VQA-RAD}~\cite{lau2018dataset} is a VQA dataset specifically designed for radiology, consisting of 315 images and 3,515 questions with 517 possible answers. The questions in VQA-RAD are categorized as either close-ended or open-ended,  depending on whether the answer choices are limited or not. 
We follow the official dataset split for our evaluation.

\noindent \textbf{SLAKE}~\cite{liu2021slake} is an English-Chinese bilingual VQA dataset composed of 642 images and 14k questions. The questions are categorized as close-ended if answer choices are limited, otherwise open-ended.
There are $224$ possible answers in total. We only use the ``English'' part, and follow the official split.

\noindent \textbf{ImageClef-VQA-2019}~\cite{ben2019vqa} is a VQA dataset constructed based on images from MedPix~\cite{bethesda2006medpix}. It comprises 4,200 radiological images accompanied by 15,292 question-answer pairs. These questions are categorized into four types: modality, plane, organ system, and abnormality. We follow the official dataset split for our evaluation.

\subsubsection{Proposed PMC-VQA Dataset}
\label{sec:benchmark}

The dataset can be used for both multiple-choice and open-ended tasks. 

\noindent \textbf{Multi-choice Answering. } 
Four candidate answers are provided for each question as the prompt.
The model is then trained to \textbf{select the correct option} among them. The accuracy (ACC) score can be used to evaluate the performance of the model on this task.

\noindent \textbf{Open-ended Answering. } 
The total possible answers for PMC-VQA are over $100$K, 
which challenges the traditional retrieval-based approach for the answer set of such a level. Therefore, we provide another training style, called ``blank'', 
where the network is not provided with options in input and is required to \textbf{directly generate answers}.
For evaluation, we adopt two metrics, Bleu scores~\cite{papineni2002bleu} and ACC scores.

We compare with strong generative models in the field of computer vision~(Open-Flamingo~\cite{awadalla2023openflamingo} and BLIP-2~\cite{li2023blip}).
Open-Flamingo~\cite{awadalla2023openflamingo} is an open-source implementation of the prior state-of-the-art generalist visual-language model, namely, Flamingo from Google DeepMind~\cite{Alayrac2022FlamingoAV},  which was trained on large-scale data from general visual-language domain.  
We utilized the released checkpoint for zero-shot evaluation in our study.
BLIP-2~\cite{li2023blip} is a pre-training strategy that bootstraps vision-language pre-training from off-the-shelf frozen pre-trained image encoders and frozen large language models. We utilized their off-shelf checkpoint for zero-shot evaluation.

\subsubsection{Pre-trained Backbones}
\label{sec:Pre-trained Models}
In this section, we introduce the pre-trained models used in our experiments. 
We separate them into language and vision backbones. Notably, while all the following models can be used in our architecture, by default, 
we use the ``PMC-LLaMA'' (or ``PMC-LLaMA-ENC'') and ``PMC-CLIP'' as backbones since they are known to be more suitable for medical data according to previous works.
The vision models are as follows.

\noindent \textbf{CLIP~\cite{radford2021learning}}: This model is trained from scratch on a dataset of 400 million image-text pairs collected from the internet with contrastive loss. We use its ``ViT-base-patch32'' version as our visual encoder with 12 transformer layers, pre-trained on natural images.

\noindent \textbf{PMC-CLIP~\cite{lin2023pmcclip}}: This model is a medical-specific visual model based on CLIP architecture, which was trained on a dataset of $1.6$ million biomedical image-text pairs collected from PubMed open-access papers using cross-modality contrastive loss. Compared to the pre-trained visual model on natural images, PMC-CLIP is specifically designed to handle medical images and text.

Our experimental approach encompasses a range of language models, enabling us to explore the pivotal role of medical knowledge and the significance of its integration into this complex task. Specifically, the language models as as follows.

\noindent \textbf{LLaMA~\cite{touvron2023llama}}: This is a state-of-the-art large-scale language model, pre-trained on trillions of tokens and widely used in the research community. We adopt the 7B version, which consists of 32 transformer layers, as our language backbone.

\noindent \textbf{PMC-LLaMA~\cite{wu2023pmcllama}}: This is an open-source language model that is acquired by fine-tuning LLaMA-7B on a total of 4.8 million biomedical academic papers with auto-regressive loss. Compared to LLaMA, PMC-LLaMA demonstrates stronger fitting capabilities and better performance on medical tasks. 

\noindent \textbf{PubMedBERT~\cite{gu2021domain}}: This is an encoder-based BERT-like model that is trained from scratch using abstracts from PubMed and full-text articles from PubMedCentral in the corpus ``The Pile''~\cite{gao2020pile}. It has 12 transformer layers and 100 million parameters. Such domain-specific models proved to yield excellent text embedding capability before the era of large language models.

\noindent \textbf{LLaMA-ENC and PMC-LLaMA-ENC.}: While LLaMA and PMC-LLaMA are known for their performance in text generation tasks, we also experiment with them as encoder models by passing a full attention mask and sampling the embedding from the last token. This allows for a direct comparison to be made with the aforementioned BERT-like models, which are also encoder-based.

\subsubsection{\changee{Implementation Details}}
\label{sec:implement}
\changee{Our models are all trained using the AdamW optimizer~\cite {loshchilov2017decoupled} with a learning rate of 2e-5.
The max context length is set as 512, and the batch size is 128.
To improve the training speed of our models,
we adopt the Deepspeed acceleration strategy, together with Automatic Mixed Precision~(AMP) and gradient checkpointing~\cite{feng2021optimal}.
All models are implemented in PyTorch and trained on 8 NVIDIA A100 GPUs with 80 GB memory.}

\subsubsection{Baseline Methods}
We compare our proposed model with established generative models (Open-Flamingo~\cite{awadalla2023openflamingo}, BLIP-2\cite{li2023blip}) and state-of-the-art approaches across various medical visual question answering models (Hanlin~\cite{ben2019vqa}, MEVF-BAN~\cite{nguyen2019overcoming}, CPRD-BAN~\cite{liu2021contrastive}, M3AE~\cite{chen2022multi}, PMC-CLIP ~\cite{lin2023pmcclip}). 

\noindent\textbf{Open-Flamingo~\cite{awadalla2023openflamingo}}: This is an open-source version of Google DeepMind's cutting-edge visual language model, Flamingo. Trained on a vast corpus of general visual-language data, Open-Flamingo represents a benchmark in the field. We utilized the released checkpoint for zero-shot evaluation in our study.

\noindent \textbf{BLIP-2~\cite{li2023blip}}: This is a robust visual-language generative model developed by Salesforce, surpassing Flamingo in reported capabilities. For our study, we utilized the released checkpoint for zero-shot evaluation. 

\noindent \textbf{Hanlin~\cite{ben2019vqa}}: This approach denotes the best overall result of the 17 participating teams in the VQA-Med 2019 task. \changee{Considering the VQA-Med 2019 dataset shares an official test split, we directly borrow the results reported in the public leaderboards\footnote{\url{https://www.aicrowd.com/challenges/imageclef-2019-vqa-med/leaderboards}}. }

\noindent \textbf{MEVF-BAN~\cite{nguyen2019overcoming}}: This approach introduces a framework that combines an unsupervised denoising auto-encoder with supervised Meta-Learning to quickly adapt to the VQA problem in scenarios with limited labeled data.
\changee{We utilize the results of MEVF-BAN on various VQA benchmarks as reported by PMC-CLIP \cite{lin2023pmcclip}, where MEVF-BAN is finetuned on each specific dataset and evaluated on the corresponding official test set.}

\noindent \textbf{CPRD-BAN~\cite{liu2021contrastive}}: This approach proposes a two-stage pre-training framework that focuses on learning transferable features from radiology images and distilling a compact visual feature extractor tailored for Med-VQA tasks. \changee{Similarly to MEVF-BAN, we adopt the results of CPRD-BAN reported in PMC-CLIP \cite{lin2023pmcclip} following the finetuning setting.}

\noindent \textbf{M3AE~\cite{chen2022multi}}: This approach is a self-supervised learning approach using multimodal masked autoencoders to learn cross-modal knowledge by reconstructing missing information from partially masked images and texts. \changee{Similarly, we adopt the results of M3AE on various MedVQA datasets as reported in PMC-CLIP \cite{lin2023pmcclip}. The official checkpoint is finetuned on each dataset and subsequently evaluated on the official test set.}

\changee{\noindent \textbf{PMC-CLIP~\cite{lin2023pmcclip}}: 
For the VQA task under zero-shot settings, we directly employed it to match image embeddings with the most similar text embeddings obtained from question-and-answer choices and then calculated the accuracy.}


\subsubsection{Evaluation Metrics}
We adopt two conventional metrics from the NLP community, 
BLEU-1 scores~\cite{papineni2002bleu}~(BiLingual Evaluation Understudy) and ACC scores~(Accuracy).

\noindent \textbf{BLEU-1.} 
BLEU-1 scores focus on the precision of unigrams, or single words, by comparing the model prediction to reference texts, yielding a score between 0 and 1.

\change{
\noindent \textbf{ACC.} 
ACC scores refer to the percentage of correctly answered questions out of the total number of questions. 
For the generative model, we calculate ACC scores by matching the model's output with the options using \texttt{difflib.SequenceMatcher}~\footnote{\url{https://docs.python.org/3/library/difflib.html}} and choosing the most similar one, which is more difficult than the evaluation for retrieval-based methods due to the unlimited output space.
Note that, \texttt{difflib.SequenceMatcher} is a class in the \texttt{difflib} module of the Python Standard Library. It is based on the Ratcliff-Obershelp algorithm, to compare sequences of elements, such as strings, lists, or any other iterable objects, and find the similarities and differences between them.}

\section{Conclusion}
In conclusion, this paper addresses the challenge of Medical Visual Question Answering~(MedVQA). Specifically, we reframe the problem of MedVQA as a generation task that naturally mirror the human-machine interactions.
We introduce a generative model for medical visual understanding
by aligning visual information from a pre-trained vision encoder with a large language model. 
To facilitate the model training,
we present a scalable pipeline for constructing PMC-VQA, a comprehensive MedVQA dataset comprising 227k VQA pairs across 149k images, spanning diverse modalities and diseases. 
Our proposed model delivers state-of-the-art performance on existing MedVQA datasets, providing a new and reliable benchmark for evaluating different methods in this field.

\section{Code Availability}
Our model checkpoint can be found in \url{https://huggingface.co/xmcmic/MedVInT-TE} and \url{https://huggingface.co/xmcmic/MedVInT-TD}, and our codes can be found in \url{https://github.com/xiaoman-zhang/PMC-VQA}.

\section{Data Availability}
The proposed dataset PMC-VQA can be found in \url{https://huggingface.co/datasets/xmcmic/PMC-VQA}. \changee{The papers used for developing PMC-VQA are from the ``Commercial Use Allowed'' split of PMC Open Access Subset\footnote{\url{https://www.ncbi.nlm.nih.gov/pmc/tools/openftlist/}}. We provide the detailed PubMed Central ID for each paper and corresponding licenses on huggingface~\footnote{\url{https://huggingface.co/datasets/xmcmic/PMC-VQA/blob/main/oa_comm_use_file_list.csv}},
which are all under CC0 or CC BY licenses. Our final dataset PMC-VQA is under CC BY-SA licenses so that it can be widely used to support the development of medical generative-based VQA models.
}
The other used public dataset can be found as follows. SLAKE is available at \url{https://www.med-vqa.com/slake/}.
VQA-RAD is available at \url{https://osf.io/89kps/}.
Image-Clef-2019 is available at  \url{https://www.imageclef.org/2019}.

\bibliographystyle{sn-mathphys} 
\bibliography{references} 

\begin{thebibliography}{10}\itemsep=-1pt

\bibitem{alayrac2022flamingo}
Jean-Baptiste Alayrac, Jeff Donahue, Pauline Luc, Antoine Miech, Iain Barr,
  Yana Hasson, Karel Lenc, Arthur Mensch, Katherine Millican, Malcolm Reynolds,
  et~al.
\newblock Flamingo: a visual language model for few-shot learning.
\newblock {\em Advances in Neural Information Processing Systems},
  35:23716--23736, 2022.

\bibitem{Alayrac2022FlamingoAV}
Jean-Baptiste Alayrac, Jeff Donahue, Pauline Luc, Antoine Miech, Iain Barr,
  Yana Hasson, Karel Lenc, Arthur Mensch, Katie Millican, Malcolm Reynolds,
  Roman Ring, Eliza Rutherford, Serkan Cabi, Tengda Han, Zhitao Gong, Sina
  Samangooei, Marianne Monteiro, Jacob Menick, Sebastian Borgeaud, Andy Brock,
  Aida Nematzadeh, Sahand Sharifzadeh, Mikolaj Binkowski, Ricardo Barreira,
  Oriol Vinyals, Andrew Zisserman, and Karen Simonyan.
\newblock Flamingo: a visual language model for few-shot learning.
\newblock In {\em Advances in Neural Information Processing Systems (NeurIPS)},
  2022.

\bibitem{antonelli2022medical}
Michela Antonelli, Annika Reinke, Spyridon Bakas, Keyvan Farahani, Annette
  Kopp-Schneider, Bennett~A Landman, Geert Litjens, Bjoern Menze, Olaf
  Ronneberger, Ronald~M Summers, et~al.
\newblock The medical segmentation decathlon.
\newblock {\em Nature Communications}, 13(1):4128, 2022.

\bibitem{awadalla2023openflamingo}
Anas Awadalla, Irena Gao, Joshua Gardner, Jack Hessel, Yusuf Hanafy, Wanrong
  Zhu, Kalyani Marathe, Yonatan Bitton, Samir Gadre, Jenia Jitsev, et~al.
\newblock Openflamingo, 2023.

\bibitem{bajwa2021artificial}
Junaid Bajwa, Usman Munir, Aditya Nori, and Bryan Williams.
\newblock Artificial intelligence in healthcare: transforming the practice of
  medicine.
\newblock {\em Future healthcare journal}, 8(2):e188--e194, 2021.

\bibitem{ben2019vqa}
Asma Ben~Abacha, Sadid~A Hasan, Vivek~V Datla, Dina Demner-Fushman, and Henning
  M{\"u}ller.
\newblock Vqa-med: Overview of the medical visual question answering task at
  imageclef 2019.
\newblock In {\em Proceedings of CLEF (Conference and Labs of the Evaluation
  Forum) 2019 Working Notes}. 9-12 September 2019, 2019.

\bibitem{ben2021overview}
Asma Ben~Abacha, Mourad Sarrouti, Dina Demner-Fushman, Sadid~A Hasan, and
  Henning M{\"u}ller.
\newblock Overview of the vqa-med task at imageclef 2021: Visual question
  answering and generation in the medical domain.
\newblock In {\em Proceedings of the CLEF 2021 Conference and Labs of the
  Evaluation Forum-working notes}. 21-24 September 2021, 2021.

\bibitem{bethesda2006medpix}
Md BETHESDA.
\newblock Medpix™ receives patent, 2006.

\bibitem{changpinyo2021conceptual}
Soravit Changpinyo, Piyush Sharma, Nan Ding, and Radu Soricut.
\newblock Conceptual 12m: Pushing web-scale image-text pre-training to
  recognize long-tail visual concepts.
\newblock In {\em Proceedings of the IEEE/CVF conference on computer vision and
  pattern recognition}, pages 3558--3568, 2021.

\bibitem{chen4578568chatffa}
Xiaolan Chen, Pusheng Xu, Yao Li, Weiyi Zhang, Fan Song, Ying-Feng Zheng, Danli
  Shi, and Mingguang He.
\newblock Chatffa: Interactive visual question answering on fundus fluorescein
  angiography image using chatgpt.
\newblock {\em Available at SSRN 4578568}.

\bibitem{chen2022multi}
Zhihong Chen, Yuhao Du, Jinpeng Hu, Yang Liu, Guanbin Li, Xiang Wan, and
  Tsung-Hui Chang.
\newblock Multi-modal masked autoencoders for medical vision-and-language
  pre-training.
\newblock In {\em Medical Image Computing and Computer Assisted Intervention},
  pages 679--689. Springer, 2022.

\bibitem{chen2024chexagent}
Zhihong Chen, Maya Varma, Jean-Benoit Delbrouck, Magdalini Paschali, Louis
  Blankemeier, Dave Van~Veen, Jeya Maria~Jose Valanarasu, Alaa Youssef,
  Joseph~Paul Cohen, Eduardo~Pontes Reis, et~al.
\newblock Chexagent: Towards a foundation model for chest x-ray interpretation.
\newblock {\em arXiv preprint arXiv:2401.12208}, 2024.

\bibitem{cheng2022dwt}
Jianhong Cheng, Hulin Kuang, Qichang Zhao, Yahui Wang, Lei Xu, Jin Liu, and
  Jianxin Wang.
\newblock Dwt-cv: Dense weight transfer-based cross validation strategy for
  model selection in biomedical data analysis.
\newblock {\em Future Generation Computer Systems}, 135:20--29, 2022.

\bibitem{vicuna2023}
Wei-Lin Chiang, Zhuohan Li, Zi Lin, Ying Sheng, Zhanghao Wu, Hao Zhang, Lianmin
  Zheng, Siyuan Zhuang, Yonghao Zhuang, Joseph~E. Gonzalez, Ion Stoica, and
  Eric~P. Xing.
\newblock Vicuna: An open-source chatbot impressing gpt-4 with 90\%* chatgpt
  quality, March 2023.

\bibitem{clusmann2023future}
Jan Clusmann, Fiona~R Kolbinger, Hannah~Sophie Muti, Zunamys~I Carrero,
  Jan-Niklas Eckardt, Narmin~Ghaffari Laleh, Chiara Maria~Lavinia L{\"o}ffler,
  Sophie-Caroline Schwarzkopf, Michaela Unger, Gregory~P Veldhuizen, et~al.
\newblock The future landscape of large language models in medicine.
\newblock {\em Communications medicine}, 3(1):141, 2023.

\bibitem{demirhan2023survey}
Hilmi Demirhan and Wlodek Zadrozny.
\newblock Survey of multimodal medical question answering.
\newblock {\em BioMedInformatics}, 4(1):50--74, 2023.

\bibitem{feng2021optimal}
Jianwei Feng and Dong Huang.
\newblock Optimal gradient checkpoint search for arbitrary computation graphs.
\newblock In {\em Proceedings of the IEEE/CVF Conference on Computer Vision and
  Pattern Recognition}, pages 11433--11442, 2021.

\bibitem{gao2020pile}
Leo Gao, Stella Biderman, Sid Black, Laurence Golding, Travis Hoppe, Charles
  Foster, Jason Phang, Horace He, Anish Thite, Noa Nabeshima, et~al.
\newblock The pile: An 800gb dataset of diverse text for language modeling.
\newblock {\em arXiv preprint arXiv:2101.00027}, 2020.

\bibitem{ge2022self}
Xiaolong Ge, Yanpeng Qu, Changjing Shang, Longzhi Yang, and Qiang Shen.
\newblock A self-adaptive discriminative autoencoder for medical applications.
\newblock {\em IEEE Transactions on Circuits and Systems for Video Technology},
  32(12):8875--8886, 2022.

\bibitem{gu2021domain}
Yu Gu, Robert Tinn, Hao Cheng, Michael Lucas, Naoto Usuyama, Xiaodong Liu,
  Tristan Naumann, Jianfeng Gao, and Hoifung Poon.
\newblock Domain-specific language model pretraining for biomedical natural
  language processing.
\newblock {\em ACM Transactions on Computing for Healthcare (HEALTH)},
  3(1):1--23, 2021.

\bibitem{he2016deep}
Kaiming He, Xiangyu Zhang, Shaoqing Ren, and Jian Sun.
\newblock Deep residual learning for image recognition.
\newblock In {\em Proceedings of the IEEE conference on computer vision and
  pattern recognition}, pages 770--778, 2016.

\bibitem{he2020pathvqa}
Xuehai He, Yichen Zhang, Luntian Mou, Eric Xing, and Pengtao Xie.
\newblock Towards visual question answering on pathology images.
\newblock pages 708--718, 2020.

\bibitem{hu2024omnimedvqa}
Yutao Hu, Tianbin Li, Quanfeng Lu, Wenqi Shao, Junjun He, Yu Qiao, and Ping
  Luo.
\newblock Omnimedvqa: A new large-scale comprehensive evaluation benchmark for
  medical lvlm.
\newblock {\em arXiv preprint arXiv:2402.09181}, 2024.

\bibitem{jin2021disease}
Di Jin, Eileen Pan, Nassim Oufattole, Wei-Hung Weng, Hanyi Fang, and Peter
  Szolovits.
\newblock What disease does this patient have? a large-scale open domain
  question answering dataset from medical exams.
\newblock {\em Applied Sciences}, 11(14):6421, 2021.

\bibitem{jones2001peir}
Kristopher~N Jones, Dwain~E Woode, Kristina Panizzi, and Peter~G Anderson.
\newblock Peir digital library: Online resources and authoring system.
\newblock In {\em Proceedings of the AMIA Symposium}, page 1075. American
  Medical Informatics Association, 2001.

\bibitem{kavur2021chaos}
A~Emre Kavur, N~Sinem Gezer, Mustafa Bar{\i}{\c{s}}, Sinem Aslan, Pierre-Henri
  Conze, Vladimir Groza, Duc~Duy Pham, Soumick Chatterjee, Philipp Ernst,
  Sava{\c{s}} {\"O}zkan, et~al.
\newblock Chaos challenge-combined (ct-mr) healthy abdominal organ
  segmentation.
\newblock {\em Medical Image Analysis}, 69:101950, 2021.

\bibitem{kung2023performance}
Tiffany~H Kung, Morgan Cheatham, Arielle Medenilla, Czarina Sillos, Lorie
  De~Leon, Camille Elepa{\~n}o, Maria Madriaga, Rimel Aggabao, Giezel
  Diaz-Candido, James Maningo, et~al.
\newblock Performance of chatgpt on usmle: Potential for ai-assisted medical
  education using large language models.
\newblock {\em PLoS digital health}, 2(2):e0000198, 2023.

\bibitem{lau2018dataset}
Jason~J Lau, Soumya Gayen, Asma Ben~Abacha, and Dina Demner-Fushman.
\newblock A dataset of clinically generated visual questions and answers about
  radiology images.
\newblock {\em Scientific data}, 5(1):1--10, 2018.

\bibitem{li2024llava}
Chunyuan Li, Cliff Wong, Sheng Zhang, Naoto Usuyama, Haotian Liu, Jianwei Yang,
  Tristan Naumann, Hoifung Poon, and Jianfeng Gao.
\newblock Llava-med: Training a large language-and-vision assistant for
  biomedicine in one day.
\newblock {\em Advances in Neural Information Processing Systems}, 36, 2024.

\bibitem{li2023blip}
Junnan Li, Dongxu Li, Silvio Savarese, and Steven Hoi.
\newblock Blip-2: Bootstrapping language-image pre-training with frozen image
  encoders and large language models.
\newblock {\em arXiv preprint arXiv:2301.12597}, 2023.

\bibitem{li2023silkie}
Lei Li, Zhihui Xie, Mukai Li, Shunian Chen, Peiyi Wang, Liang Chen, Yazheng
  Yang, Benyou Wang, and Lingpeng Kong.
\newblock Silkie: Preference distillation for large visual language models.
\newblock {\em arXiv preprint arXiv:2312.10665}, 2023.

\bibitem{lin2023pmcclip}
Weixiong Lin, Ziheng Zhao, Xiaoman Zhang, Chaoyi Wu, Ya Zhang, Yanfeng Wang,
  and Weidi Xie.
\newblock Pmc-clip: Contrastive language-image pre-training using biomedical
  documents.
\newblock 2023.

\bibitem{lin2022medical}
Zhihong Lin, Donghao Zhang, Qingyi Tac, Danli Shi, Gholamreza Haffari, Qi Wu,
  Mingguang He, and Zongyuan Ge.
\newblock Medical visual question answering: A survey.
\newblock {\em arXiv preprint arXiv:2111.10056}, 2022.

\bibitem{liu2021contrastive}
Bo Liu, Li-Ming Zhan, and Xiao-Ming Wu.
\newblock Contrastive pre-training and representation distillation for medical
  visual question answering based on radiology images.
\newblock In {\em Medical Image Computing and Computer Assisted Intervention},
  pages 210--220. Springer, 2021.

\bibitem{liu2021slake}
Bo Liu, Li-Ming Zhan, Li Xu, Lin Ma, Yan Yang, and Xiao-Ming Wu.
\newblock Slake: A semantically-labeled knowledge-enhanced dataset for medical
  visual question answering.
\newblock In {\em 2021 IEEE 18th International Symposium on Biomedical Imaging
  (ISBI)}, pages 1650--1654. IEEE, 2021.

\bibitem{liu2023qilin}
Junling Liu, Ziming Wang, Qichen Ye, Dading Chong, Peilin Zhou, and Yining Hua.
\newblock Qilin-med-vl: Towards chinese large vision-language model for general
  healthcare.
\newblock {\em arXiv preprint arXiv:2310.17956}, 2023.

\bibitem{loshchilov2017decoupled}
Ilya Loshchilov and Frank Hutter.
\newblock Decoupled weight decay regularization.
\newblock {\em arXiv preprint arXiv:1711.05101}, 2017.

\bibitem{lu2023mathvista}
Pan Lu, Hritik Bansal, Tony Xia, Jiacheng Liu, Chunyuan Li, Hannaneh
  Hajishirzi, Hao Cheng, Kai-Wei Chang, Michel Galley, and Jianfeng Gao.
\newblock Mathvista: Evaluating mathematical reasoning of foundation models in
  visual contexts.
\newblock {\em arXiv preprint arXiv:2310.02255}, 2023.

\bibitem{moor2023med}
Michael Moor, Qian Huang, Shirley Wu, Michihiro Yasunaga, Yash Dalmia, Jure
  Leskovec, Cyril Zakka, Eduardo~Pontes Reis, and Pranav Rajpurkar.
\newblock Med-flamingo: a multimodal medical few-shot learner.
\newblock In {\em Machine Learning for Health (ML4H)}, pages 353--367. PMLR,
  2023.

\bibitem{nguyen2019overcoming}
Binh~D Nguyen, Thanh-Toan Do, Binh~X Nguyen, Tuong Do, Erman Tjiputra, and
  Quang~D Tran.
\newblock Overcoming data limitation in medical visual question answering.
\newblock In {\em Medical Image Computing and Computer Assisted Intervention},
  pages 522--530. Springer, 2019.

\bibitem{nicolson2023concise}
Aaron Nicolson, Jason Dowling, and Bevan Koopman.
\newblock A concise model for medical image captioning.
\newblock In {\em CLEF (Working Notes)}, pages 1611--1619, 2023.

\bibitem{nori2023capabilities}
Harsha Nori, Nicholas King, Scott~Mayer McKinney, Dean Carignan, and Eric
  Horvitz.
\newblock Capabilities of gpt-4 on medical challenge problems.
\newblock {\em arXiv preprint arXiv:2303.13375}, 2023.

\bibitem{openai2023gpt4}
OpenAI.
\newblock Gpt-4 technical report.
\newblock {\em arXiv preprint arXiv:2303.08774}, 2023.

\bibitem{papineni2002bleu}
Kishore Papineni, Salim Roukos, Todd Ward, and Wei-Jing Zhu.
\newblock Bleu: a method for automatic evaluation of machine translation.
\newblock In {\em Proceedings of the 40th annual meeting of the Association for
  Computational Linguistics}, pages 311--318, 2002.

\bibitem{park2024patient}
Jiwoo Park, Kangrok Oh, Kyunghwa Han, and Young~Han Lee.
\newblock Patient-centered radiology reports with generative artificial
  intelligence: adding value to radiology reporting.
\newblock {\em Scientific Reports}, 14(1):13218, 2024.

\bibitem{pelka2018roco}
Obioma Pelka, Sven Koitka, Johannes R{\"u}ckert, Felix Nensa, and Christoph~M
  Friedrich.
\newblock Radiology objects in context (roco): a multimodal image dataset.
\newblock In {\em MICCAI Workshop on Large-scale Annotation of Biomedical Data
  and Expert Label Synthesis (LABELS) 2018}, pages 180--189. Springer, 2018.

\bibitem{radford2021learning}
Alec Radford, Jong~Wook Kim, Chris Hallacy, Aditya Ramesh, Gabriel Goh,
  Sandhini Agarwal, Girish Sastry, Amanda Askell, Pamela Mishkin, Jack Clark,
  et~al.
\newblock Learning transferable visual models from natural language
  supervision.
\newblock In {\em International conference on machine learning}, pages
  8748--8763. PMLR, 2021.

\bibitem{roberts2001pubmed}
Richard~J Roberts.
\newblock Pubmed central: The genbank of the published literature.
\newblock volume~98, pages 381--382. National Acad Sciences, 2001.

\bibitem{safranek2023role}
Conrad~W Safranek, Anne~Elizabeth Sidamon-Eristoff, Aidan Gilson, and David
  Chartash.
\newblock The role of large language models in medical education: applications
  and implications, 2023.

\bibitem{seyfioglu2023quilt}
Mehmet~Saygin Seyfioglu, Wisdom~O Ikezogwo, Fatemeh Ghezloo, Ranjay Krishna,
  and Linda Shapiro.
\newblock Quilt-llava: Visual instruction tuning by extracting localized
  narratives from open-source histopathology videos.
\newblock {\em arXiv preprint arXiv:2312.04746}, 2023.

\bibitem{singhal2022large}
Karan Singhal, Shekoofeh Azizi, Tao Tu, S~Sara Mahdavi, Jason Wei, Hyung~Won
  Chung, Nathan Scales, Ajay Tanwani, Heather Cole-Lewis, Stephen Pfohl, et~al.
\newblock Large language models encode clinical knowledge.
\newblock {\em arXiv preprint arXiv:2212.13138}, 2022.

\bibitem{subramanian-2020-medicat}
Sanjay Subramanian et~al.
\newblock Medicat: A dataset of medical images, captions, and textual
  references.
\newblock In {\em Findings of EMNLP}, 2020.

\bibitem{thirunavukarasu2023large}
Arun~James Thirunavukarasu, Darren Shu~Jeng Ting, Kabilan Elangovan, Laura
  Gutierrez, Ting~Fang Tan, and Daniel Shu~Wei Ting.
\newblock Large language models in medicine.
\newblock {\em Nature medicine}, 29(8):1930--1940, 2023.

\bibitem{touvron2023llama}
Hugo Touvron, Thibaut Lavril, Gautier Izacard, Xavier Martinet, Marie-Anne
  Lachaux, Timoth{\'e}e Lacroix, Baptiste Rozi{\`e}re, Naman Goyal, Eric
  Hambro, Faisal Azhar, et~al.
\newblock Llama: Open and efficient foundation language models.
\newblock {\em arXiv preprint arXiv:2302.13971}, 2023.

\bibitem{wang2023chatcad}
Sheng Wang, Zihao Zhao, Xi Ouyang, Qian Wang, and Dinggang Shen.
\newblock Chatcad: Interactive computer-aided diagnosis on medical image using
  large language models.
\newblock {\em arXiv preprint arXiv:2302.07257}, 2023.

\bibitem{wang2017chestx}
Xiaosong Wang, Yifan Peng, Le Lu, Zhiyong Lu, Mohammadhadi Bagheri, and
  Ronald~M Summers.
\newblock Chestx-ray8: Hospital-scale chest x-ray database and benchmarks on
  weakly-supervised classification and localization of common thorax diseases.
\newblock In {\em Proceedings of the IEEE conference on computer vision and
  pattern recognition}, pages 2097--2106, 2017.

\bibitem{wu2023pmcllama}
Chaoyi Wu, Xiaoman Zhang, Ya Zhang, Yanfeng Wang, and Weidi Xie.
\newblock Pmc-llama: Towards building open-source language models for medicine.
\newblock {\em arXiv preprint arXiv:2304.14454}, 2023.

\bibitem{wu2023towards}
Chaoyi Wu, Xiaoman Zhang, Ya Zhang, Yanfeng Wang, and Weidi Xie.
\newblock Towards generalist foundation model for radiology.
\newblock {\em arXiv preprint arXiv:2308.02463}, 2023.

\bibitem{wu2024hallucination}
Jinge Wu, Yunsoo Kim, and Honghan Wu.
\newblock Hallucination benchmark in medical visual question answering.
\newblock {\em arXiv preprint arXiv:2401.05827}, 2024.

\bibitem{yang2023impact}
Jiancheng Yang, Hongwei~Bran Li, and Donglai Wei.
\newblock The impact of chatgpt and llms on medical imaging stakeholders:
  perspectives and use cases.
\newblock {\em Meta-Radiology}, page 100007, 2023.

\bibitem{yang2021medmnist}
Jiancheng Yang, Rui Shi, and Bingbing Ni.
\newblock Medmnist classification decathlon: A lightweight automl benchmark for
  medical image analysis.
\newblock In {\em 2021 IEEE 18th International Symposium on Biomedical Imaging
  (ISBI)}, pages 191--195. IEEE, 2021.

\bibitem{zhan2023unidcp}
Chenlu Zhan, Yufei Zhang, Yu Lin, Gaoang Wang, and Hongwei Wang.
\newblock Unidcp: Unifying multiple medical vision-language tasks via dynamic
  cross-modal learnable prompts.
\newblock {\em arXiv preprint arXiv:2312.11171}, 2023.

\bibitem{Zhang2022OPTOP}
Susan Zhang, Stephen Roller, Naman Goyal, Mikel Artetxe, Moya Chen, Shuohui
  Chen, Christopher Dewan, Mona Diab, Xian Li, Xi~Victoria Lin, Todor Mihaylov,
  Myle Ott, Sam Shleifer, Kurt Shuster, Daniel Simig, Punit~Singh Koura, Anjali
  Sridhar, Tianlu Wang, and Luke Zettlemoyer.
\newblock Opt: Open pre-trained transformer language models.
\newblock {\em arXiv preprint arXiv:2205.01068}, 2022.

\bibitem{zhang2023biomedclip}
Sheng Zhang, Yanbo Xu, Naoto Usuyama, Hanwen Xu, Jaspreet Bagga, Robert Tinn,
  Sam Preston, Rajesh Rao, Mu Wei, Naveen Valluri, et~al.
\newblock Biomedclip: a multimodal biomedical foundation model pretrained from
  fifteen million scientific image-text pairs.
\newblock {\em arXiv preprint arXiv:2303.00915}, 2023.

\end{thebibliography}

\clearpage
\appendix
\section{Supplemental Materials}

\captionsetup[figure]{name=Supplementary Fig.}
\captionsetup[table]{name=Supplementary Table }

\subsection{Data Analysis}
\label{supple:data_analysis}
\begin{figure}[htb]
    \centering
    \includegraphics[width=0.9\textwidth]{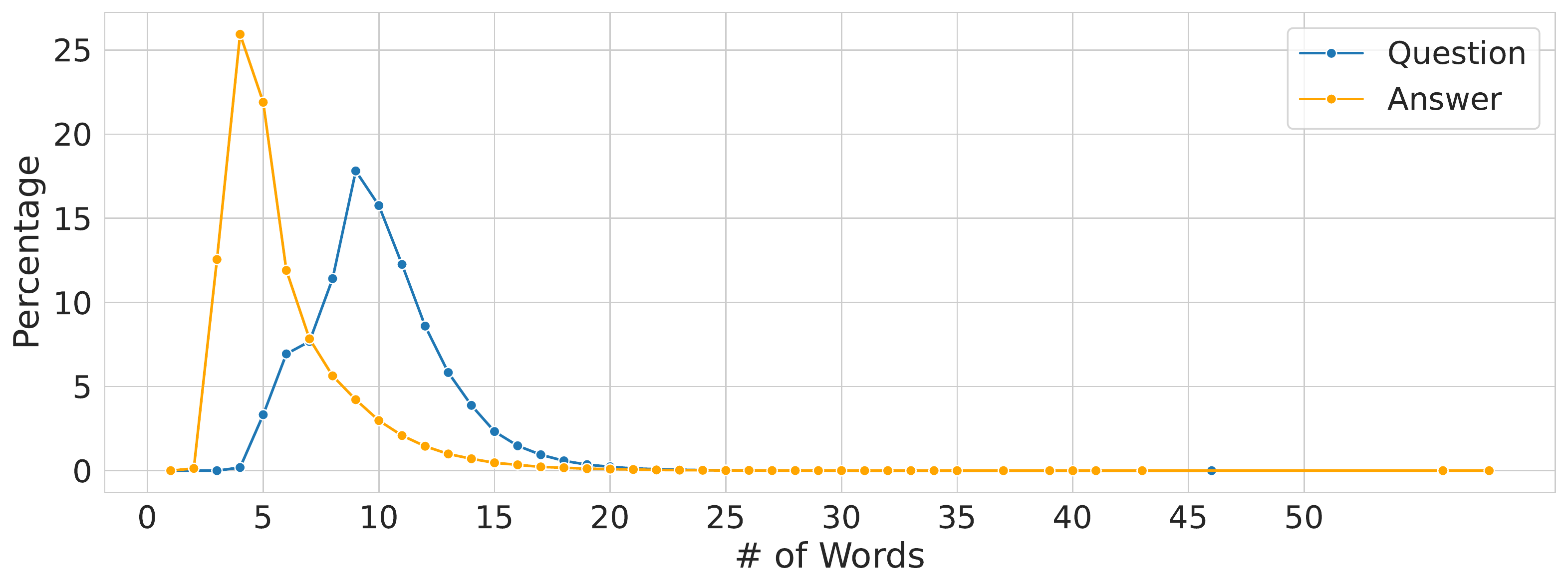}
    \caption{Percentage of questions and answers with different word lengths.}
    \label{fig:question_length}
\end{figure}

Fig.~\ref{fig:question_length} shows the percentage of questions and answers with different word lengths.
Most questions range from 5 to 15 words, and most answers are around 5 words.

\subsection{Evaluation on Original Split Test Set}
\label{supple:eval_on_original}
In this section, we report the experimental results on the original randomly split test set in Supplementary Table~\ref{tab:pmcvqa_originaltest}.
This test set is more extensive but did not undergo additional manual verification. As indicated, the performance experienced a slight decline when compared with the PMC-VQA-test, yet the reduction was minimal. For instance, the accuracy (ACC) for the choice task decreased from 40.3 to 39.2. This slight variation underscores the inherent high quality and robustness of our dataset.

\begin{table}[!tbh]
\centering
\scriptsize 
\setlength{\tabcolsep}{6pt}
\caption{\change{Comparison of baseline models using different pre-trained models on both open-ended and multiple-choice tasks. We reported the results on \changeee{PMC-VQA-test-initial}. `Scratch' means to train the vision model from scratch with the same architecture as `PMC-CLIP'.}
}
\renewcommand\arraystretch{1.05}
\begin{tabular}{lllccc}
\toprule
\multirow{2}{*}{\textbf{Method}}                                    & \multirow{2}{*}{\textbf{Language Backbone}}          & \multirow{2}{*}{Vision Backbone}                      & {Choice}         & \multicolumn{2}{c}{Blanking}                  \\ \cmidrule{4-6} 
                                                                    &                                                      &                                                       & ACC    & BLEU-1 & ACC   \\  \midrule
\multicolumn{5}{l}{\textbf{Zero-shot}}\\ \midrule
PMC-CLIP~\cite{lin2023pmcclip}                   & PMC-CLIP~\cite{lin2023pmcclip}    & PMC-CLIP~\cite{lin2023pmcclip}     & {24.0 (23.4,24.6)}         & - & -          \\  
BLIP-2~\cite{li2023blip}                               & OPT-2.7B~\cite{Zhang2022OPTOP}    & CLIP~\cite{radford2021learning}    & {24.6 (23.9,25.2)}     & {22.5 (21.9,23.2)}              & {5.2 (4.8,5.7)}                     \\  
Open-Flamingo~\cite{awadalla2023openflamingo}    & LLaMA~\cite{touvron2023llama}           & CLIP~\cite{radford2021learning}    & {25.0 (24.5, 25.6)}    & {26.1 (25.6,26.7)}              & {4.1 (3.7, 4.6)}                      \\ 
\changee{LLaVA-Med~\cite{li2024llava}} & \changee{Vicuna~\cite{vicuna2023}} & \changee{BioMedCLIP~\cite{zhang2023biomedclip}} & \changee{32.6 (32.1,33.2)} & \changee{28.3 (27.8,28.8)} & \changee{3.7(3.7,3.8)}  \\
\midrule
\multicolumn{5}{l}{\textbf{Trained on PMC-VQA}}\\ \midrule
LLaMA~\cite{touvron2023llama}                    & LLaMA~\cite{touvron2023llama}     & --                                                    & {30.6 (30.0,31.2)}       & {26.1 (25.7,26.8)}              & {14.2 (13.9,14.6)}            \\ \midrule
\multirow{9}{*}{MedVInT-TE}                                         & \multirow{3}{*}{PubMedBERT~\cite{gu2021domain}}        & Scratch                                        & {34.4 (33.7,35.1)}   & {33.7 (33.0, 34.5)} & \underline{20.4 (19.9,20.9)} \\  
                                                                    &                                                      & CLIP~\cite{radford2021learning}    & {34.5 (33.8,35.1)} &  {33.7 (32.9,34.3)}  & \underline{20.4 (20.0,20.9)}  \\  
                                                                    &                                                      & PMC-CLIP~\cite{lin2023pmcclip}     & {37.1 (36.4,37.9)}   & \underline{35.2 (34.6,36.0)}  & \textbf{22.0 21.6,22.4)}  \\ \cmidrule{2-6} 
                                                                    & \multirow{3}{*}{LLaMA-ENC~\cite{touvron2023llama}}     & Scratch                                               & {35.2 (34.5,35.9)}     & {32.5 (31.7,33.1)}  & {15.3 (14.8,15.7)}    \\  
                                                                    &                                                      & CLIP~\cite{radford2021learning}    & {35.3 (34.7,35.9)}    &{32.3 (31.5,33.0)}  & {15.6 (14.8,15.7)}      \\   
                                                                    &                                                      & PMC-CLIP~\cite{lin2023pmcclip}     & {36.9 (36.2,37.6)}   & \textbf{35.4 (34.8,36.1)}  & {18.2 (17.7,18.6)} \\ \cmidrule{2-6} 
                                                                    & \multirow{3}{*}{PMC-LLaMA-ENC~\cite{wu2023pmcllama}}   & Scratch                                               & {37.0(36.3,37.6)}     & {32.6 (32.0,33.3)}  & {16.2 (15.7,16.6)}      \\   
                                                                    &                                                      & CLIP~\cite{radford2021learning}    & {37.1(36.4,37.9)}   & {33.0 (32.1,33.7)}  & {16.6 (16.2,17.0)}     \\  
                                                                    &                                                      & PMC-CLIP~\cite{lin2023pmcclip}     & \underline{38.2 (37.5,38.9)}    & {34.8 (34.0,35.3)}  & {18.1 (17.7,18.6)}        \\  \midrule
\multirow{6}{*}{MedVInT-TD}                                         & \multirow{3}{*}{LLaMA~\cite{touvron2023llama}}               & Scratch                                               & {36.2 (35.7,36.9)}          & {29.1 (28.1,29.7)}              & {17.4 (17.2,17.9)}              \\   
                                                                    &                                                      & CLIP~\cite{radford2021learning}    & \underline{38.2 (37.5,38.9)}    & {31.3 (30.6,32.0)}              & {19.5 (19.1,20.0)}                \\  
                                                                    &                                                      & PMC-CLIP~\cite{lin2023pmcclip}     & {37.3 (36.8,38.0)}       & {31.9 (31.2,32.6)}              & {20.0 (19.6,20.5)}       \\ \cmidrule{2-6} 
                                                                    & \multirow{3}{*}{PMC-LLaMA~\cite{wu2023pmcllama}}      & Scratch                                               & {36.8 (36.2,37.6)}        & {28.6 (27.8,29.1)}              & {16.8 (16.4,17.1)}           \\  
                                                                    &                                                      & CLIP~\cite{radford2021learning}    & {36.8 (36.2,37.5)}    & {31.4 (30.8,32.1)}              & {19.5 (19.1,20.0)}        \\ 
                                                                    &                                                      & PMC-CLIP~\cite{lin2023pmcclip}  & \textbf{39.4 (38.7,40.0)}     & {32.7 (31.1,33.2)}              & {20.3 (19.9,20.7)}            \\   \bottomrule
\end{tabular}
\label{tab:pmcvqa_originaltest}
\end{table}

\subsection{Ablation Study}
\label{supple:ablation_study}
In this section, we add the comparison of baseline models using different projection modules~(MLP or Transformer) on both open-ended and multiple-choice tasks.
\changee{
MLP-based projection module, employs a two-layer Multilayer Perceptron (MLP), while the second variant, transformer-based projection modules, employs a 12-layer transformer decoder supplemented with several learnable vectors as query input.}
As shown in Table~\ref{tab:ablation}, different projection modules demonstrate comparable performance across various evaluation tasks.
Both architectures can effectively reconcile the diversity in the embedding dimensions arising from different pre-trained visual models, making our architecture adaptable to various visual foundation model designs, regardless of whether they are based on VIT or ResNet.

\begin{table}[htb]
\centering
\footnotesize
\setlength{\tabcolsep}{3pt}
\caption{Ablation study of baseline models using different projection modules and pre-trained models on open-ended and multiple-choice tasks. We reported the results of the original test set of the PMC-VQA/PMC-VQA test. ``Scratch'' means to train the vision model from scratch with the same architecture as ``PMC-CLIP''.} 
\vspace{3pt}
\begin{tabular}{l|l|l|cc|c}
\toprule
\multirow{2}{*}{Method} & \multirow{2}{*}{Language Backbone} & \multirow{2}{*}{Vision Backbone} & \multicolumn{2}{c|}{Blanking} & Choice    \\ 
 & & & ACC  & Bleu-1   & ACC  \\ \midrule
\multirow{9}{*}{MedVInT-TE-MLP} & \multirow{3}{*}{PubMedBERT~\cite{gu2021domain}} & Scratch  & 33.7 / 34.2 & {20.4} / 20.9 & 34.4 / 34.9\\
 & & CLIP~\cite{radford2021learning}    & 32.3 / 34.4 & 15.6 / 20.8 & 34.5 / 34.3\\
 & & PMC-CLIP~\cite{lin2023pmcclip}  & {35.2} / {36.4} & {\bf 22.0} / {\bf 23.2} & 37.1 / 37.6\\ \cmidrule{2-6} 
 & \multirow{3}{*}{LLaMA-ENC~\cite{touvron2023llama}} & Scratch  & 32.5  /  32.5 & 15.3 / 15.9 & 35.2 / 35.2\\
 & & CLIP~\cite{radford2021learning}     & 32.3 / 33.4 & 15.6 / 15.1 & 35.3 / 36.1\\
 & & PMC-CLIP~\cite{lin2023pmcclip}  & {35.4} / {\bf 36.8} & 18.2 / 18.4 & 36.9 / 37.1 \\ \cmidrule{2-6} 
 & \multirow{3}{*}{PMC-LLaMA-ENC~\cite{wu2023pmcllama}}      & Scratch  & 32.6 / 35.0 & 16.2 / 17.0 & 37.0 / 38.0 \\
 & & CLIP~\cite{radford2021learning}     & 33.0 / 34.4  & 16.6 / 16.5 & 37.1 / 38.5 \\
 & & PMC-CLIP~\cite{lin2023pmcclip}  & 34.8  / 35.3 & 18.1 / 18.6 & {38.2} / 39.2 \\ \midrule
 \multirow{9}{*}{MedVInT-TE-Transformer} & \multirow{3}{*}{PubMedBERT~\cite{gu2021domain}} & Scratch  & 34.1 / 36.2 & {\bf 21.0} / {\bf 21.9} &  39.8 / 40.6\\
 & & CLIP~\cite{radford2021learning}    & 33.9 / 34.6  & 20.6 / 21.8 & {\bf 39.9} / 40.9   \\
 & & PMC-CLIP~\cite{lin2023pmcclip}  & 33.7 / 35.4 &  20.3 / 21.2 & {\bf 40.2} / 40.9   \\ \cmidrule{2-6} 
 & \multirow{3}{*}{LLaMA-ENC~\cite{touvron2023llama}} & Scratch  &  32.0 / 33.5 & 15.1 / 15.3 & 38.4 / 39.7 \\
 & & CLIP~\cite{radford2021learning}     & 32.3 / 34.3  & 15.5 / 15.7 & 38.4 / 38.7  \\
 & & PMC-CLIP~\cite{lin2023pmcclip}  & {\bf 35.9} / {\bf 37.1} & 19.0 / 19.3 & 38.9 / 39.4 \\ \cmidrule{2-6} 
 & \multirow{3}{*}{PMC-LLaMA-ENC~\cite{wu2023pmcllama}}   & Scratch & 33.2 / 34.7 & 16.6 / 16.5 & 38.1 /39.8 \\
 & & CLIP~\cite{radford2021learning}     & 33.6 / 35.1 & 16.7 / 17.2 & 38.7 / 38.9 \\
 & & PMC-CLIP~\cite{lin2023pmcclip}  & {\bf 35.5} / 36.0 & 18.4 /18.6 & 38.2 / 37.7 \\ 
 \midrule
\multirow{6}{*}{MedVInT-TD-MLP}     & \multirow{3}{*}{LLaMA\cite{touvron2023llama}} & Scratch  &  28.1 / 30.6 & 16.5 / 16.9 & 35.8 / 37.4 \\
 & & CLIP~\cite{radford2021learning}     &  30.2 / 32.7  & 18.6 / 18.5 & 35.8 / 37.1  \\
 & & PMC-CLIP~\cite{lin2023pmcclip}  &    31.3 / 32.6 &  19.5 / 19.8 &  38.4 / {\bf 41.0}  \\ \cmidrule{2-6}
 & \multirow{3}{*}{PMC-LLaMA~\cite{wu2023pmcllama}}      & Scratch  &   28.3 / 30.6 & 16.4 / 17.3  & 35.8 / 37.0 \\
 & & CLIP~\cite{radford2021learning}     &   31.4 / 31.8 & 19.2 / 19.5 & 36.2 / 37.9\\
 & & PMC-CLIP~\cite{lin2023pmcclip}  &   32.1 / 31.7 & 19.7 / 20.2 &  38.4 / {\bf 42.3}\\ \midrule
 \multirow{6}{*}{MedVInT-TD-Transformer}     & \multirow{3}{*}{LLaMA\cite{touvron2023llama}} & Scratch  &29.1 / 30.2 & 17.4 / 18.0 & 36.2 / 37.9\\
 & & CLIP~\cite{radford2021learning}     &  31.3 / 32.2 & 19.5 / 20.0 & 38.2 / 39.2\\
 & & PMC-CLIP~\cite{lin2023pmcclip}  & 31.9 / 33.4 & 20.0 /  21.3 & 37.3 / {39.5}\\ \cmidrule{2-6}
 & \multirow{3}{*}{PMC-LLaMA~\cite{wu2023pmcllama}}      & Scratch  & 28.6 / 29.8 &  16.8 / 17.4 & 36.8 / 36.9\\
 & & CLIP~\cite{radford2021learning}     & 31.4 / 32.6 & 19.5 / 20.4 & 36.8 / 36.9\\
 & & PMC-CLIP~\cite{lin2023pmcclip}  & 32.7 / 33.6 & 20.3  / {21.5} & {39.4 / 40.3}\\
 \bottomrule
\end{tabular}
\label{tab:ablation}
\end{table}

\subsection{Fail Case Study}
\begin{figure}[tbh]
    \centering
    \includegraphics[width=0.95\textwidth]{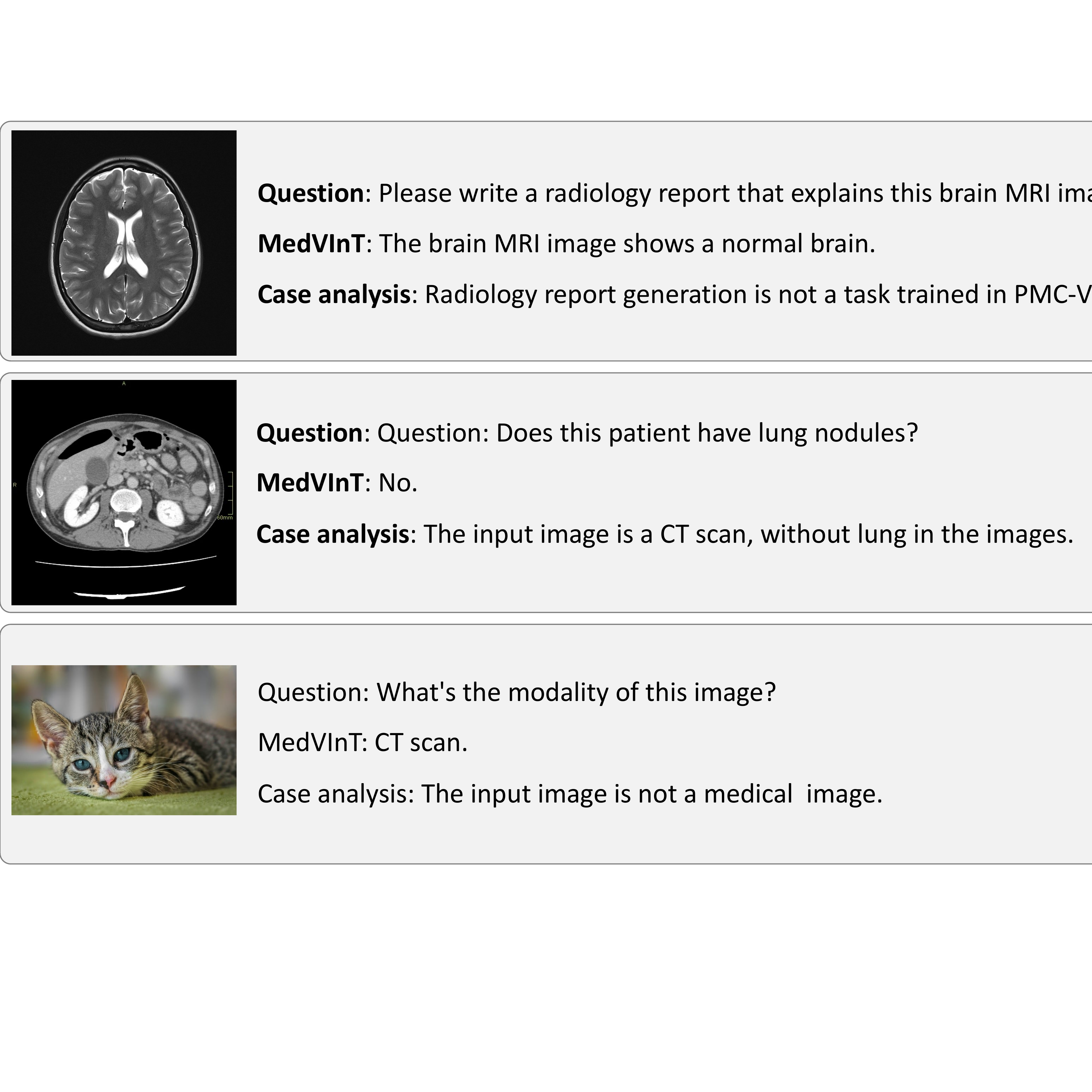}
    \caption{\changee{Examples of ``out of scope'' and  ``non-sensical'' questions.}
    }
\label{fig:fail_casestudy}
\end{figure}

\changee{
In this section, we explore the hallucinations exhibited by the proposed MedVInT models. As a starting point for generative-based MedVQA methods, for now, our models still suffer from hallucinations in nonsensical or adversarial cases with huge domain gaps.
As illustrated in Fig~\ref{fig:fail_casestudy}, 
for out-of-scope tasks such as report generation, 
the model may not produce radiology reports in a structured format. However, it sometimes provides reasonable answers. 
For nonsensical questions, such as inquiring about lung nodules in an abdomen CT image, the model cannot refuse to answer nor highlight the mistake in the question. 
}

\end{document}